\documentclass[journal, letterpaper]{IEEEtran}

\usepackage{graphicx}

\usepackage{url}

\usepackage{textgreek}	
\usepackage{listings}
\usepackage{csvsimple}
\usepackage{longtable}

\usepackage{amsmath}
\usepackage{amsthm}
\usepackage{amssymb}
\usepackage[utf8]{inputenc} 
\usepackage[T1]{fontenc}    
\usepackage{nicefrac}       
\usepackage{microtype}      
\usepackage{xcolor}         
\usepackage{kotex}
\usepackage{wrapfig}
\usepackage{mathtools}
\usepackage{floatrow} 
\newfloatcommand{capbtabbox}{table}[][\FBwidth]  

\usepackage{url}
\usepackage{graphicx}
\usepackage{booktabs}
\usepackage{tikz}
\usepackage{wrapfig}
\usepackage{subcaption}

\usepackage{multicol}

\usepackage{grffile}
\usepackage{epstopdf}
\AppendGraphicsExtensions{.tif}

\usepackage{algorithm, algpseudocode} 
\usepackage{mathrsfs}

\begin{document}

        \title{ All-In-One: Artificial Association Neural Networks}
        \author{Seokjun Kim$^{1}$,
        Jaeeun Jang$^{2}$
        and~Hyeoncheol Kim$^{*}$\\
       Department of Computer Science and Engineering\\
            Korea University\\
            \{melon7607, wkdwodms0779, harrykim\}@korea.ac.kr\\%
        }
	\maketitle

\begin{abstract}
Most deep learning models are limited to specific datasets or tasks because of network structures using fixed layers. In this paper, we discuss the differences between existing neural networks and real human neurons, propose association networks to connect existing models, and describe multiple types of deep learning exercises performed using a single structure. Further, we propose a new neural data structure that can express all basic models of existing neural networks in a tree structure. We also propose an approach in which information propagates from leaf to a root node using the proposed recursive convolution approach (i.e., depth-first convolution) and feed-forward propagation is performed. Thus, we design a ``data-based,'' as opposed to a ``model-based,'' neural network. In experiments conducted, we compared the learning performances of the models specializing in specific domains with those of models simultaneously learning various domains using an association network. The model learned well without significant performance degradation compared to that for models performing individual learning. In addition, the performance results were similar to those of the special case models; the output of the tree contained all information from the tree. Finally, we developed a theory for using arbitrary input data and learning all data simultaneously.
\end{abstract}

\section{Introduction}
	\IEEEPARstart{T}{hus} far, various deep learning models have been reported in the literature, e.g., multilayer perceptron(MLP), convolutional, recurrent, recursive, and graph neural networks\cite{ lecun1989backpropagation,hopfield1982neural, scarselli2008graph, wu2020comprehensive}, and these different networks achieve good performances in various fields.
Most previous studies focus on specific datasets or tasks because there are many tasks that only humans can do and neural networks cannot.
\begin{figure*}[!t]
\centering
\hspace{-20px}
\subfloat[Human brain]{{\includegraphics[width=0.45\textwidth]{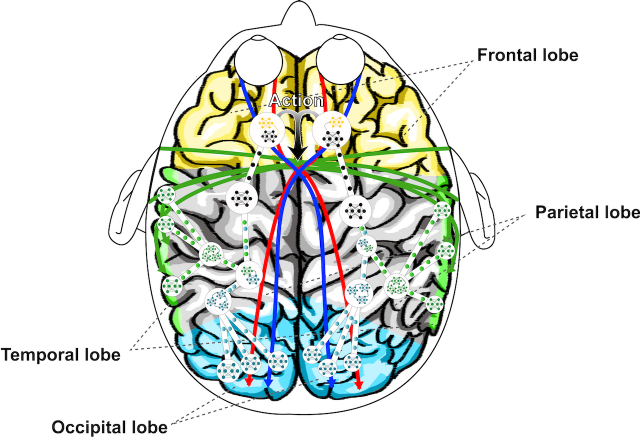} }}
\subfloat[Artificial association networks]{{\includegraphics[width=0.30\textwidth]{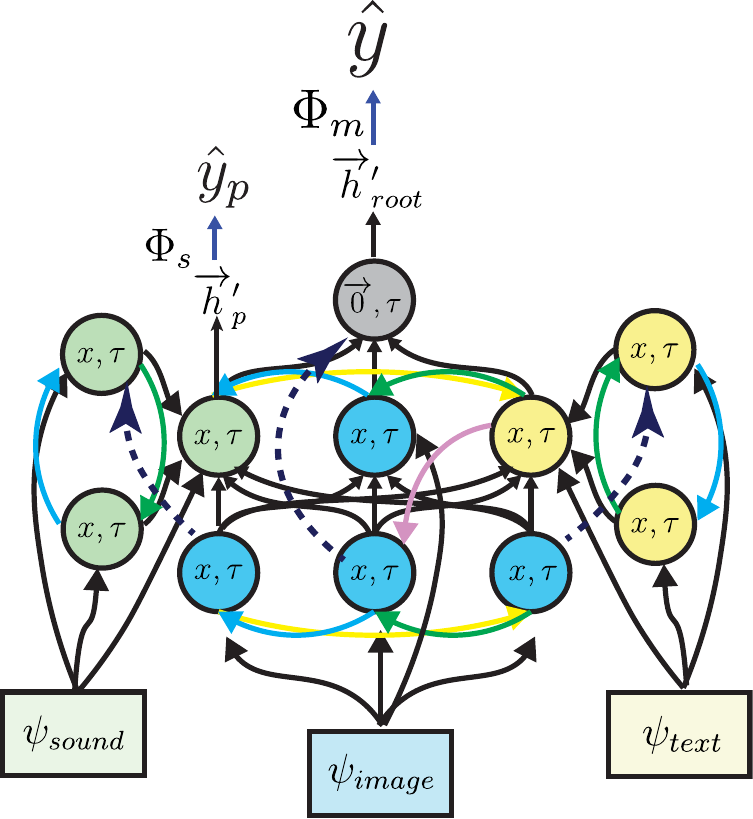}}}
\caption{Comparison of the human brain with an artificial association neural network.}
\label{fig:network_comparison}
\end{figure*}

We analyze human neural network structures and conduct research to imitate this structure and investigating the discrepancy.
First, we describe some difficulties encountered when imitating the human neural network structure, then we summarize our approach towards these problems for developing the artificial association networks (AANs) proposed in this study.

Hereafter, let C-$[\cdot]$ denote the characteristics of the human brain and A-$[\cdot]$ denote our approaches to emulating C-$[\cdot]$.

\textbf{(C-\romannumeral 1)} Various sensory organs process information, and there are various locations where this information is received in the cerebral cortex.
Sensory organs provide humans with visual, auditory, and olfactory information, and this information is transferred to the cerebral cortex using different information-processing organs.
Further, the optic nerve starts with the visual cortex of the occipital lobe, and the auditory nerve with the auditory cortex of the temporal lobe; thus, the two paths have different starting points.

\textbf{(A-\romannumeral 1)} We define each sensory organ as a feature extraction process to solve the abovementioned problem.
Feature extraction networks for each domain are stored in a dictionary data structure for creating a multifeature extraction model (Section \ref{sec:multi_feature_extraction}).
Each type of data requires a different feature extraction process based on the domain, and they each have a domain bias.

\textbf{(C-\romannumeral 2)} The human brain does not independently process various information domains. Instead, the information received from other sensory organs is gathered in the association area\cite{pandya1982association}.

When we receive information, we process it in addition to the information of various domains around us.
For example, if we hear the \textbf{voice domain} information on the word ``coffee,'' we build a relationship between ``latte'' in the \textbf{text domain} and ``mug'' in the \textbf{image domain}. 
This implies that information processing from different domains is not independent. 
For example, sensory, visual, and auditory information are fused in the somatosensory, visual, and auditory association areas, respectively. In addition, several types of information are associated with each other. The posterior parietal cortex\cite{malach1995object} is located at the top of the head as part of the parietal lobe, wherein visual and auditory information are fused and interpreted; further, this information is coordinated in the frontal lobe and humans act accordingly.

This structure inspired us; all information has various relationships, and we plan to develop a data structure such as a graph (relational) or tree (hierarchical) to express these relationships.
However, a tree has only one parent node, and it cannot express the relationship between sibling nodes.
Furthermore, it is difficult to define the starting and ending points when using a graph. 
Finally, the hidden state of the t-th layer cannot be used at layer t+2 because the hidden state is updated at layer t+1.

\textbf{(A-\romannumeral 2)} We propose two new data structures: neuronode and neurotree.
A neuronode represents an entity that can express the characteristics of the graph and tree data structures. It has connections with sibling nodes in a tree structure and can have multiple parent nodes.
We consider the relationships between entities and save them as neuronodes; they are used to create a neurotree.

This structure can be seen as a graph; however, it is a feed-forward structure that cannot be reversed. Thus, the neurotree includes directional information from leaf nodes to root nodes, and this path is traversed using recursive propagation\cite{goller1996learning}.

\textbf{(C-\romannumeral 3)} The propagation path of the brain is overly complex. In addition, the characteristics of the brain change continuously \cite{sagi2012learning, opendak2015adult}; the brain has various paths, and it processes information from multiple domains within a single structure.
However, most existing models are structurally fixed and designed such that it limits them to specific tasks or datasets. 
Studies on routing networks\cite{rosenbaum2019routing} have already discussed these issues and mentioned problems with fixed neural network configurations and module selectors.

Therefore, it is problematic to express neural network layers in a fixed structure.
The neural networks need to express various structures based on the information being processed.
In contrast, most existing models are structurally fixed, and therefore, they are limited to specific tasks or datasets.

\textbf{(A-\romannumeral 3)} We do not design layers; however, we express the flow of information delivery using a neurotree (Section \ref{sec:neurotree_architecture}).
We extract features as in (A-\romannumeral 1) and build a neurotree as in (A-\romannumeral 2).
 
\textbf{(C-\romannumeral 4)} The number of neurons and processing depth determine the complexity that neurons can process. In addition, the number of neurons used differs based on the domain and complexity of the data. 
The higher the number of neurons, the greater is the complexity of the information that can be processed.
However, neurons are pruned as humans get older, which helps to eliminate unnecessary connections\cite{petanjek2011extraordinary}.

Here, we consider that the number of convolutions and connections should be flexible.
Further, deep neural networks exhibit a related overfitting problem\cite{widrow1960adaptive} that occurs when using deeper layers\cite{dai2017very}; various attempts have been made to solve this using dropout\cite{srivastava2014dropout}.
NasNet\cite{zoph2018learning} attempted to automatically design a layer through reinforcement learning and recurrent neural networks (RNNs); thus,  there is an appropriate network depth based on the complexity of the dataset.

\textbf{(A-\romannumeral 4)} The height of the neurotree in (A-\romannumeral 3) indicates the number of convolutions. 
We plan to design a network that can adjust the layer depth based on the complexity of each item of data and not that of the dataset, instead of using a fixed depth.

\textbf{(C-\romannumeral 5)} There are a significant number of human neurons and they process various data domains. In addition, the human brain contains approximately 85 billion neurons\cite{herculano2009human}, and this quantity is difficult to express in a network.

\textbf{(A-\romannumeral 5)} The AANs can train multiple neurons using a single network cell.
We propose a  depth-first convolution (DFC) methodology (Section \ref{sec:Depthfirst}), which is a recursive propagation methodology in which AANs can learn all information in the mini-batch neurotree.

This network theory states that units of information have a relationship in the form of a graph, which then becomes a larger unit of information and has a relationship with other units of information. Meanwhile, the unit of information is a set of neurons, and we express it as a vector with artificial association networks.

This study was conducted to develop connecting models that can combine various neural networks for implementing the human brain as a function of $f(\exists x) \rightarrow y$ that can simultaneously learn for $\forall x$. 
This is a neural network model that can use any information from a real-world environment as input.

This approach proposes data structures (neuronode, neurotree) that can represent existing neural networks in trees; furthermore, we propose an artificial association neural network that learns these structures.

 With this model, it is possible to express complex information structures in neurotree and perform subtasks simultaneously. Most data structures (e.g., tabular, image, sound, text, deep sequence, graph, and tree data) can be expressed as a neurotree. Thus, this model can be a structurally free, data-driven, and domain-general all-in-one model.
The rest of this paper is organized as follows. Section 2 provides an overview of related work. Section 3 discusses the development of artificial association networks, and this can be applied to multidomain deep learning. Section 4 describes experiments performed to validate the performance of the proposed network. Finally, Section 5 provides a discussion of its merits and concludes this paper.

\section{Related Work}
\subsection{Graph Convolutional Networks}
Graph neural networks (GNNs) have demonstrated good performance in various fields\cite{scarselli2008graph, wu2020comprehensive, velivckovic2017graph, gilmer2017neural}. 
GNNs have a relational term and a structure that is more similar to human neural networks compared to other networks.
Graph convolutional networks (GCNs)\cite{kipf2016semi} do not consider multidimensional edge features; however, they are simple and easy to use, as indicated in 
\begin{equation}
\label{eq:gcn1}
\mathbf{h}'_{t} = f(\mathbf{A}, \mathbf{h}_{t})
\end{equation}
Equation (\ref{eq:gcn1}) illustrates the basic principle of this type of network. Here, $t$, $\mathbf{A}$, $\mathbf{h}_{t}$, and $N$ denote the order of the layer, adjacency matrix for relationship information, hidden states for node features, and node size, respectively. We can express this term as $\mathbf{h}_{t} = \{\overrightarrow{h}_{t1}, \overrightarrow{h}_{t2}... \overrightarrow{h}_{tN}\}$. The input is $\mathbf{h}_{0}$ as $\mathbf{x}$, and $\mathbf{h}'_{t}$ denotes the outputs.
\begin{equation}
\label{eq:gcn2}
\mathbf{h}'_{t} = \sigma(\mathbf{A}\mathbf{h}_{t}\mathbf{W}_{t}^{T})
\end{equation}
This process expresses (1) in more detail;
$\mathbf{A}\in\mathbb{R}^{N \times N}$, where $N$ represents the number of nodes.
Further, $F$, $F'$, $\mathbf{W}\in\mathbb{R}^{F' \times F}$, $\mathbf{h}_{t}\in\mathbb{R}^{N \times F}$, and $\mathbf{h}'_{t}\in\mathbb{R}^{N \times F'}$ represent 
the input size, output size, weight parameters, 
hidden state, and output, respectively.
\begin{equation}
\label{eq:gcn3}
\mathbf{h}'_{t} = \sigma(\tilde{\mathbf{D}}^{-\frac{1}{2}}\tilde{\mathbf{A}}\tilde{\mathbf{D}}^{-\frac{1}{2}}\mathbf{h}_{t}\mathbf{W}_{t}^{T})
\end{equation}
In this study, the renormalization trick is applied to utilize the spectral graph convolution for deep learning.
$\tilde{\mathbf{A}}=\mathbf{I}+\mathbf{A}$, where $\mathbf{I}$ and $\tilde{\mathbf{D}}$ represent the identity matrix and order matrix of $\tilde{\mathbf{A}}$, respectively.
Through this process, $\mathbf{A}$ represents normalized and relational information that can be learned.
Finally, the overall information is aggregated through a readout process, which results in the output.

We can implement an equivalent graph neural network (EGNN)(C)\cite{gong2019exploiting}, which allows us to learn multidimensional edge features by slightly modifying this model.

\subsection{Graph Attention Networks}
\begin{equation}
\label{eq:gat}
\alpha_{tij} = \frac{\mathbf{exp}(LeakyReLU(\overrightarrow{\mathbf{a}}_{t}^{T}[\mathbf{W}_{t}\overrightarrow{h}_{ti},\mathbf{W}_{t}\overrightarrow{h}_{tj}]))}{\sum_{k\in \mathcal{N}_{i}}\mathbf{exp}(LeakyReLU(\overrightarrow{\mathbf{a}}_{t}^{T}[\mathbf{W}_{t}\overrightarrow{h}_{ti},\mathbf{W}_{t}\overrightarrow{h}_{tk}]))}
\end{equation}
Graph attention networks (GATs)\cite{velivckovic2017graph} use a normalized attention score by applying an attention mechanism to $\mathbf{A}$.
A parameter $\overrightarrow{\mathbf{a}}_{t}^{T}$ is added for the attention mechanism $\mathbb{R}^{2F'} \times \mathbb{R}^{2F'} \rightarrow \mathbb{R}$, and LeakyReLU is used as the attention's activation function\cite{xu2015empirical}.
$\mathcal{N}_{i}$ represents a set of nodes connected to the node of the i-th child in $\mathbf{A}$.
\begin{equation}
\label{eqn:gat2}
\mathbf{h}'_{t} = \sigma((\mathbf{A}\odot\mathbf{\alpha}_{t})\mathbf{h}_{t}\mathbf{W}_{t}^{T})
\end{equation}
$\odot$ denotes the hadamard product, which is applied with $\alpha_{t}$ to the adjacency matrix in (\ref{eq:gcn2}) to form (\ref{eqn:gat2}); this  model learns how the i-th node is of importance to the j-th node.

We implement EGNN(A)\cite{gong2019exploiting}, and further enable the learning of multidimensional edge features by slightly modifying this model.

\subsection{Message-Passing Neural Networks}
Message-passing neural networks (MPNNs) deal with graph data structures\cite{gilmer2017neural}. Unlike the original GCN and GATs, this type of neural network can handle multidimensional edge features.

Each node in the graph uses a message-passing technique that receives information from neighbors, and it repeatedly updates information of the current node.
\begin{equation}
\label{eq:mpnn1}
m^{t+1}_{i} = \sum_{j \in N(i)}M_{t}(h^{t}_{i}, h^{t}_{j}, e_{ij})
\end{equation}
\begin{equation}
\label{eq:mpnn2}
h^{t+1}_{i} = U_{t}(h^{t}_{i}, m^{t+1}_{i})
\end{equation}
The order of the message delivery process performed T times is expressed as t.
Neighboring nodes connected to node $i$ of graph G are called $N(i)$; $j$ denotes a node connected to $i$, and $j \in N(i)$.
In this case, the hidden state of node $i$ is called $h_{i}$, and the hidden state of node $j$ is called $h_{j}$; the edge feature to which $i$ and $j$ are connected is called $e_{ij}$.
In the t-th step, the message function is called $M_{t}$, and the function that updates the vertex is $U_{t}$. Therefore, the above process can be written as follows.
\begin{equation}
\label{eq:mpnn3}
h^{t+1}_{i} = U_{t}(h^{t}_{i}, \sum_{j \in N(i)}M_{t}(h^{t}_{i}, h^{t}_{j}, e_{ij}))
\end{equation}
\begin{equation}
\label{eq:mpnn4}
\hat{y} = g(\{h^{t}_{i}|i \in G\})
\end{equation}
Finally, the overall information is aggregated ($g$) through a readout process, and the result is the output, as shown in (\ref{eq:mpnn4}).

\subsection{Recursive Neural Networks}
Recursive neural networks (RecNNs) are useful in semantics-related fields such as natural language processing\cite{socher-etal-2013-recursive}.
We believe that RecNNs cannot be used effectively because they require tree-structured data.
RecNNs can learn hierarchical information in trees and feed-forward delivery structures from leaf nodes to the root node, and they can easily understand propagation paths.

\subsection{Tree-based Convolutional Neural Networks}
An abstract syntax tree\footnote{https://docs.python.org/3/library/ast.html} (AST) is a data structure used to express program codes in compilers.

The source code is written in a programming language such as C, Java, Python, or Go; the computer converts this code into machine language and executes instructions based on the grammar thereof.
An AST represents grammar and instructions in a tree structure, and each AST node contains diverse types of information such as for and while loops, function definitions and calls, and assign and load operations.

In studies on tree-based convolutional neural networks (TBCNNs)\cite{mou2014tbcnn, bui2018cross}, the source code is converted into an AST, and the types of each AST node are converted into a continuous vector space through representation learning. Then, information is embedded using convolution and pooling layers.
In this learning process, similar types of instructions, such as for and while loops, are represented with similar vectors. This process has experimentally demonstrated good performance.
\subsection{Control Flow Graphs}
A control flow graph (CFG)\cite{allen1970control} is a graph representation that is useful in static analysis for programming the source code. This approach allows us to express an execution path of code that can be executed during the execution of the program.
The nodes of a CFG are called basic blocks; a basic block contains code statements that run sequentially without jumping, and the edges of the CFG indicate jumps in the control flow. In addition, there are entry blocks at the starting point and exit blocks at the ending point.

CFGs are effective when used to train GNNs \cite{8809504}.
\subsection{Depth-First Search} 
\label{sec:related_works_dfs} 
\begin{figure}[ht]
\centering
\begin{subfigure}{0.49\textwidth}
    \centering
    \includegraphics[height=2.0cm]{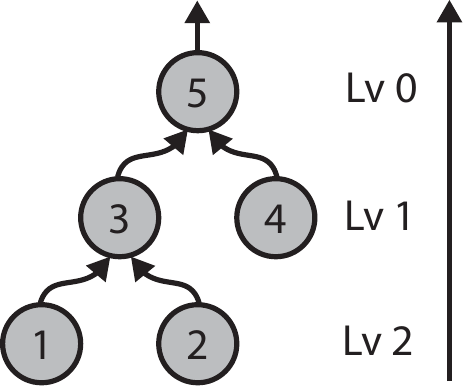}
    \caption{Post-order and left-side-first}
    \label{fig:dfs_post}
\end{subfigure}
\hfill
\begin{subfigure}{0.49\textwidth}
    \centering
    \includegraphics[height=2.0cm]{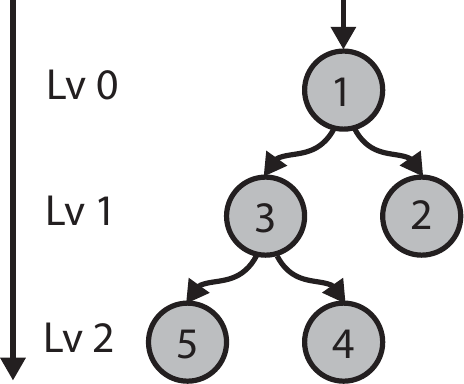}
    \caption{Pre-order and right-side-first}
    \label{fig:dfs_pre}
\end{subfigure}
\caption{Two types of depth-first search algorithms for tree navigation.}
\label{fig:dfs_algorithm}
\end{figure}
The depth-first search (DFS) algorithm traverses the data in a tree or graph structure. In this study, the DFS algorithm is used to explore tree data structures.
We search in the order shown in Fig. \ref{fig:dfs_post}, Appendix \ref{algorithm:dfs} when we travel in the post-order and left-side-first order.
We use pre-order and right-side-first search and travel in the order shown in Fig. \ref{fig:dfs_pre}, Appendix \ref{algorithm:dfs} to tour post-order and left-side-first search methods in the reverse order.
The DFS order is suitable as a propagation methodology in RecNNs that learn tree data structures.

We slightly modify this algorithm and propose an algorithm that learns the neurotree (Section \ref{sec:Depthfirst}).
\subsection{Multi Deep Learning}

Multidomain deep learning entails learning datasets of different domains using a single network.
One related study designed parallel residual adapters\cite{rebuffi2018efficient}, which learn several types of image data. In this study, 10 different domain image datasets (including MNIST, CIFAR-100, and VGG-Flower) were used to learn one neural network structure. In particular, this approach can be highly effective in reducing parameters, as only one model is used for learning on datasets.

Multimodal deep learning involves combining data from various domains and solving the task using the fused information. An image captioning field\cite{yu2019multimodal} is a representative example.

Multi-task deep learning involves performing various tasks with a single neural network.
This approach has an advantage in that the characteristics of the data are more diverse.
As a representative example, the MT-DNN\cite{liu2019multi} model has shown a good performance.

\section{Artificial Association Networks}
\label{sec:artificial_association_networks}
\begin{figure}[h!]
\centering
\begin{subfigure}{0.95\textwidth}
    \centering
    \includegraphics[height=2.0cm]{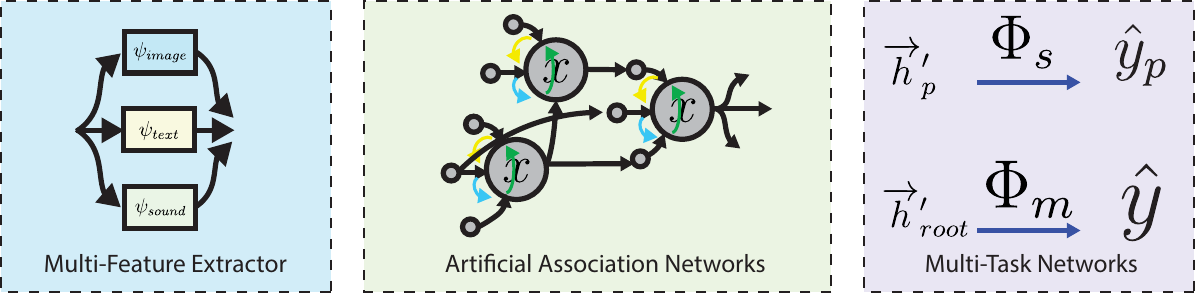}
\end{subfigure}
\caption{Three stages of association model.}
\label{fig:3-stage-neurotree}
\end{figure}
Humans perform multiple types of deep learning simultaneously.
A deep learning model should not be structurally fixed to perform multiple types of deep learning (multidomain, multimodal, and multi-task learning) using few parameters for all data. In this study, we introduce AANs that perform multidomain deep learning; these networks can be expanded with multimodal deep learning. Finally, in future studies, multi-task learning will be performed in the root node.

Our goal is to design a network connection model for expressing the structures of various neural network models using trees and learn through recursive convolution to explain the human brain in a single $brain(\exists x) \rightarrow y$ model.
\subsection{ Multifeature Extraction Networks and Domain Bias }
\label{sec:multi_feature_extraction}
\begin{figure}[h!]
\centering
\begin{subfigure}{0.42\textwidth}
    \centering
    \includegraphics[height=2.7cm]{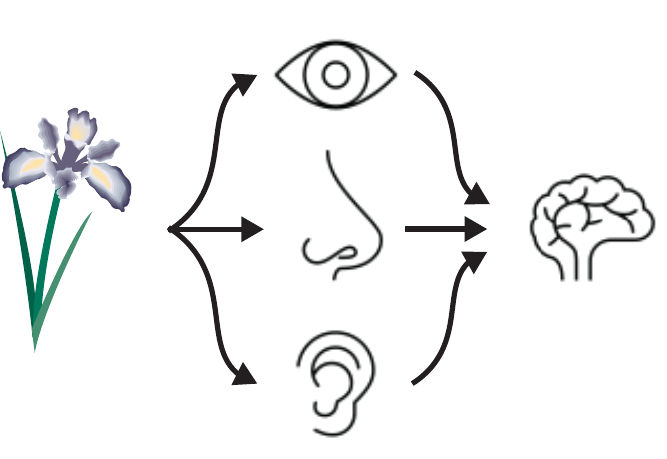}
    \caption{ Sensory nervous system }
\end{subfigure}
\hfill
\begin{subfigure}{0.55\textwidth}
    \centering
    \includegraphics[height=2.7cm]{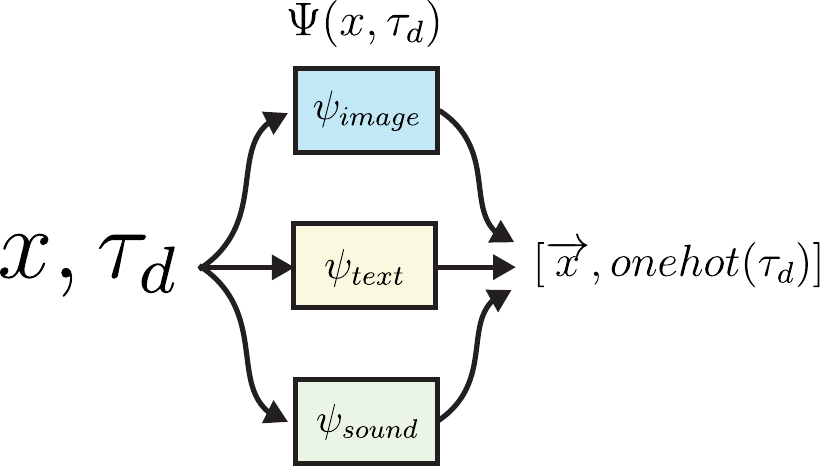}
    \caption{Multifeature extraction networks}
\end{subfigure}
\caption{Comparison of the human sensory nervous system and AAN models}
\label{fig:multi-feature-extractor-with-human}
\end{figure}
\begin{equation}
\label{eq:feature_extraction1}
\sigma(\mathbf{W}\overrightarrow{x} + b) = \overrightarrow{h}
\end{equation}
\begin{equation}
\overrightarrow{x}\in\mathbb{R}^{F\times1}, \mathbf{W}\in\mathbb{R}^{F' \times F}, b\in\mathbb{R}^{F' \times 1}, \overrightarrow{h}\in\mathbb{R}^{F'\times1}
\end{equation}
We define the process of calculating a fully connected layer(FC layer) as (\ref{eq:feature_extraction1}).
Here, $\overrightarrow{x}$, $\mathbf{W}$, $b$, and $\sigma$ denote the input vector with $F$ dimensions, weight parameters, bias, and activation function, respectively. The activation function is used to activate $W\overrightarrow{x}+b$ only if it exceeds the threshold. At this point, bias affects the threshold.
The output of the hidden state is denoted as $\overrightarrow{h}$ with $F'$ dimensions.
\begin{equation}
\label{eq:feature_extraction2}
\sigma(\mathbf{W}[\overrightarrow{x},1]) = \overrightarrow{h}
\end{equation}
\begin{equation}
[\overrightarrow{x},1]\in\mathbb{R}^{(F+1)\times1}, \mathbf{W}\in\mathbb{R}^{F' \times (F+1)}, \overrightarrow{h}\in\mathbb{R}^{F'\times1}
\end{equation}
We simplify (\ref{eq:feature_extraction1}) as (\ref{eq:feature_extraction2}); further, 
[,] denotes the concatenation. The same formula can be expressed without adding the bias by concatenating 1 to $\overrightarrow{x}$.
Here, $[\overrightarrow{x},1]$ denotes an input and bias input, and $\mathbf{W}$ denotes weight parameters with a bias value. The weight parameter corresponding to 1 becomes the bias value.
\begin{equation}
\label{eq:feature_extraction3}
\sigma(\mathbf{W}[\overrightarrow{x},onehot(\tau_{d})]) = \overrightarrow{h}
\end{equation}
\begin{equation}
[\overrightarrow{x},onehot(\tau_{d})]\in\mathbb{R}^{(F+B)\times1}, W\in\mathbb{R}^{F' \times (F+B)}, \overrightarrow{h}\in\mathbb{R}^{F' \times 1}
\end{equation}
We replace the bias value of 1 (in (\ref{eq:feature_extraction2})) with a one-hot vector for the domain (in (\ref{eq:feature_extraction3})).
$\tau_{d}$ represents domain information such as images, sound, and text; this allows each domain to learn the bias value differently.

In (\ref{eq:feature_extraction3}), weight parameters corresponding to the domain indicate the bias for the domain.
The activation threshold can be adapted for each domain.
\begin{equation}
\label{eq:feature_extraction4}
\psi_{\tau_{d}}(x) = \overrightarrow{x}
\end{equation}
\begin{equation}
\label{eq:feature_extraction5}
\sigma(W[\psi_{\tau_{d}}(x),onehot(\tau_{d})]) = \overrightarrow{h}
\end{equation}
Next, we apply $\psi_{\tau_{d}}$, which allows $\exists{x}$ to be converted to $\overrightarrow{x}$.
Meanwhile, $\psi_{\tau_{d}}$ represents a feature extraction network that processes data for domain ($\tau_{d}$), and it is an abstract representation of a sensory organ that corresponds to the domain.
\begin{equation}
\label{eq:multi_feature_extraction}
\Psi = \{\psi_{image}, \psi_{sound},\psi_{text} ..\}
\end{equation}
\begin{equation}
\label{eq:feature_extraction}
\therefore \overrightarrow{x}' = \Psi(x, \tau_{d}) = [\psi_{\tau_{d}}(x),onehot(\tau_{d})]
\end{equation}
Finally, we applied the above process (\ref{eq:feature_extraction5}) with various domains.
We create a multifeature extraction function that corresponds to the domain-specific $\psi_{\tau_{d}}$ using the dictionary data structure; this is denoted as $\Psi$. This structure is scalable, independent, and suitable for transfer learning and fine-tuning.

Therefore, these multifeature extraction networks ($\Psi$) use $\exists{x}$ and $\tau_{d}$ as inputs and it converts them into $\overrightarrow{x}'$. Further, we can express $\overrightarrow{x}$ as a zero vector ($\overrightarrow{0}$) if the neuronode is an empty node.
We need a data structure that can contain this information ($x, \tau_{d}$); to this end, we use a neuronode.

\subsection{Neural Data Structure: Neuronode and Neurotree}
\label{sec:neuro_datastructure}
\begin{figure}[h!]
\centering

\begin{subfigure}{0.90\textwidth}
    \centering
    \includegraphics[height=2.5cm]{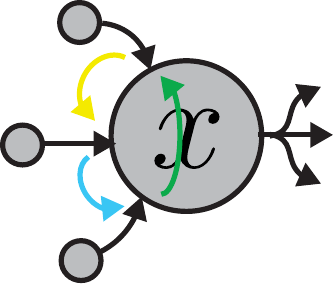}
\end{subfigure}
\caption{ Relational (graph) + hierarchical (tree) + multiple parents = Neuronode.}
\label{fig:neurotree-datastructure}
\end{figure}
Each node of the neurotree represents an entity as a set of neurons; thus, it is possible to express the relationships between and the hierarchy among various domain information.
All neuronodes can perform subtasks; the root node performs the main task.
\begin{itemize}
\item[$x:$] (Input) -- Input data on the current neuronode. This can be any type of data ($\exists x$).
For example, it is possible to store images, sounds, text, or tabular data.
\item[$\tau_{d}:$] (Domain) -- 
Domain information of the current neuronode.
In (\ref{eq:feature_extraction}), $\Psi$ receives this $\tau_{d}$ (domain) information and a feature extraction network($\psi_{\tau_{d}}$) is selected.
\item[$\tau_{s}:$] (Subtask) -- 
A subtask can be performed in each neuronode; if this information is not present, this neuronode does not perform subtasks (Section \ref{sec:multi_task}).
\item[$\mathbf{A}_c:$](Children adjacency matrix) -- 
Relationships used for aggregating information among child neuronodes. The number of children remains the same as the number of nodes. If we define the number of children as $N$, we can express this matrix as $\mathbf{A}_{c}\in\mathbb{R}^{N \times N}$.
\item[$\mathbf{C}:$] (Children) -- 
This information refers to the children of the neuronode, which can represent hierarchical information such as nested sets.
Each child corresponds to $\mathbf{A}_{c}$.
We can express this information as $\mathbf{C}=\{\mathbf{NN}_{1},\mathbf{NN}_{2},..\mathbf{NN}_{N}\}$.
\end{itemize}
This can be expressed as $\mathbf{NN} = \{x,\tau_{d}, \tau_{s},\mathbf{A}_c,\mathbf{C}\}$, where $\mathbf{NN}_{root}$ denotes the root node of $\mathbf{NT}$.

However, we cannot use $\mathbf{A}_{c}$ to learn multidimensional edge features. Instead, we can replace $\mathbf{A}_{c}$ with $\mathbf{E}_{c}$. Multidimensional edge features are not discussed in this paper; however, they will be discussed in our next study. 
There is only one parent node when making a tree.
However, neuronodes can have multiple parent nodes. That is, a neuronode can be a child ($\mathbf{C}$) of several nodes.
Here, there is one condition: ``the descendants of a node cannot include the ancestors of the node.'' This condition prevents cycles from occurring within the hierarchy of the neurotree.

This structure may appear like the graph data structure; the difference is that it maintains a tree structure. In a neurotree, the links to the incoming and outgoing directions are expressed as child and multiple parent nodes, respectively; this represents a directed graph. Finally, the undirected graph is represented by $\mathbf{A}_{c}$ for aggregating input entities.
Therefore, this structure has a direction and can express a feed-forward structure that does not reverse.

The reasons for maintaining multiple parents and feed-forward propagation without considering trees in which cycles exist is as follows:
(1) The GNNs update the hidden state of every node each time a convolution is performed; however, the AANs do not update the hidden states of nodes.
Therefore, it is possible to use the t-hidden state when determining t+2 by keeping the existing hidden state in a neuronode.
In this model, the level of the neurotree indicates the number of layers, and it is possible to transfer hidden states by jumping between layers in the direction of propagation.
(2) It is necessary to express the number of convolutions using the height of the neurotree. In some GNN models, the number of convolutions is appropriately fixed; this information is received from neighboring nodes.
However, the recursive propagation method, i.e., DFC (Section \ref{sec:Depthfirst}), maintains a feed-forward structure. The height of the neurotree represents the number of times that various pieces of information are combined. However, the cycle must stop after the appropriate number of iterations for preventing an infinite convolution.
Thus, the cycle must be represented by $\mathbf{A}_{c}$ or removed by copying that node or not traversing the child node at the appropriate moment if a cyclical structure is required.

This data structure is defined in this manner to express the forms of various existing neural network models (multilayer, recurrent, graph, and recursive neural networks) in a tree structure; it becomes possible to learn the datasets of the existing time-series, tree, and graph structures without major modifications. Further, because of the multiple-parent structure, we can use the t-th hidden state information at t+2.
\begin{figure*}[ht]
\centering
\begin{subfigure}{0.13\textwidth}
    \centering
    \includegraphics[height=2.5cm]{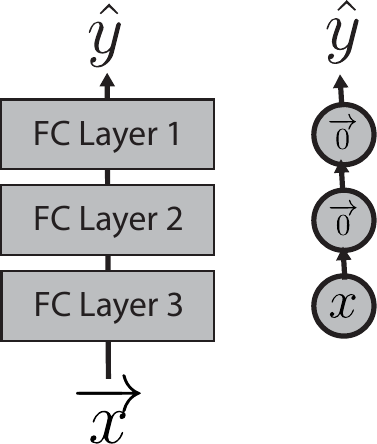}
    \caption{MLP and NT}    \label{fig:mlp-nt}

\end{subfigure}
\hfill
\begin{subfigure}{0.35\textwidth}
    \centering
    \includegraphics[height=2.5cm]{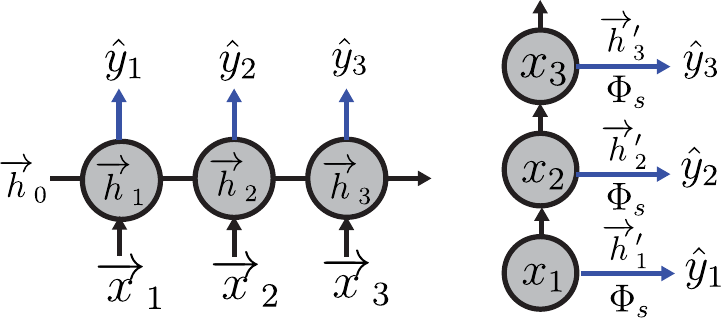}
    \caption{RNN and NT}     \label{fig:rnn-nt}

\end{subfigure}
\hfill
\begin{subfigure}{0.20\textwidth}
    \centering
    \includegraphics[height=2.5cm]{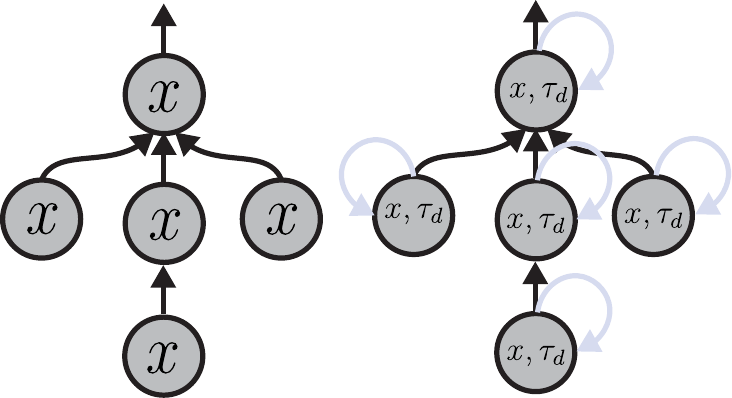}
    \caption{RecNN and NT}    \label{fig:recnn-nt}

\end{subfigure}
\hfill
\begin{subfigure}{0.20\textwidth}
    \centering
    \includegraphics[height=2.5cm]{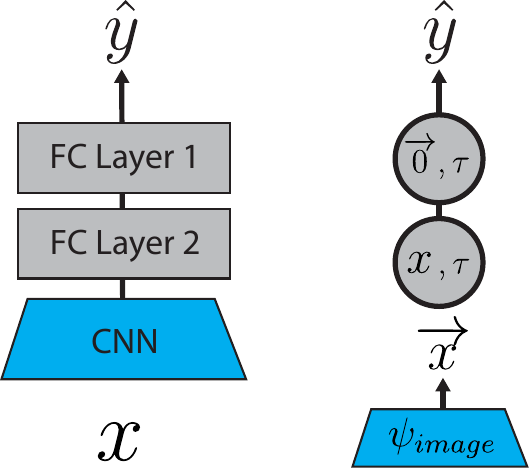}
    \caption{CNN and NT}     \label{fig:cnn-nt}

\end{subfigure}
\hfill
\begin{subfigure}{0.33\textwidth}
    \centering
    \includegraphics[height=3.0cm]{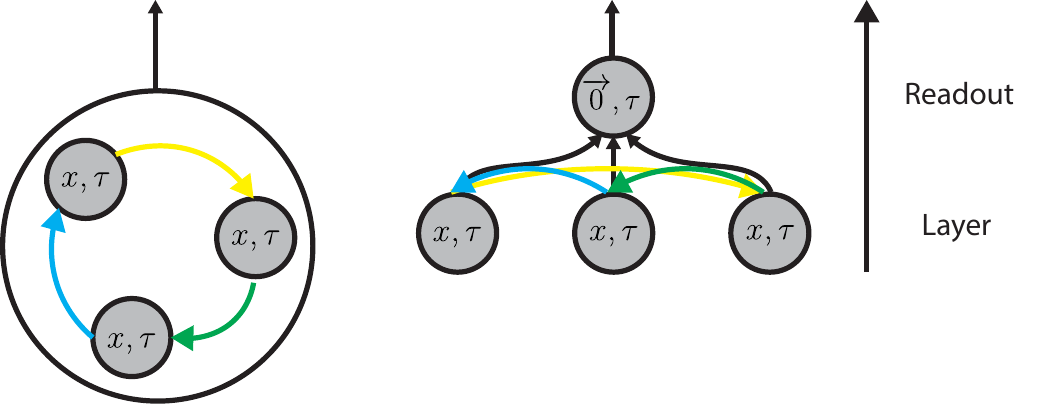}
    \caption{GNN and NT}     \label{fig:gnn-nt}

\end{subfigure}
\hfill
\begin{subfigure}{0.26\textwidth}
    \centering
    \includegraphics[height=3.0cm]{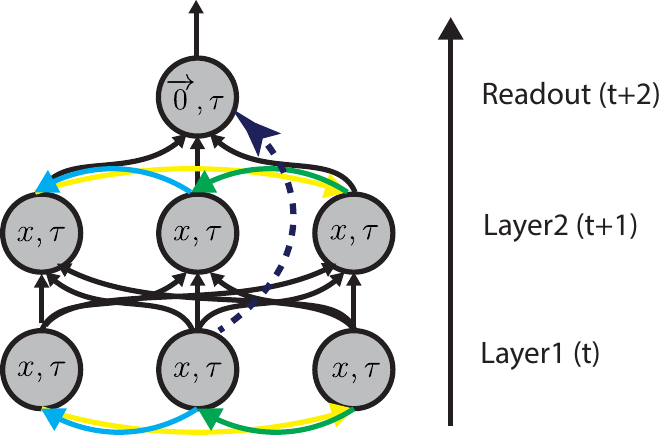}
    \caption{ Multiparent graph neural networks }
    \label{fig:multi-parent-gnn}
\end{subfigure}
\hfill
\begin{subfigure}{0.22\textwidth}
    \centering
    \includegraphics[height=3.0cm]{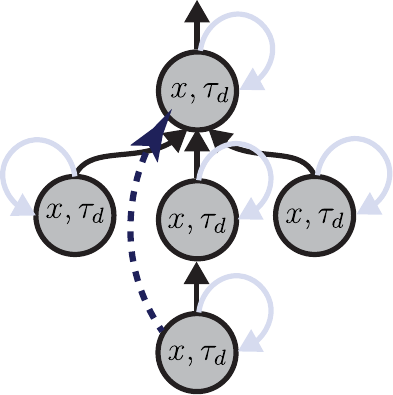}
    \caption{ Multiparent recursive neural networks }
    \label{fig:multi-parent-recursive-networks}
\end{subfigure}
\caption{Model structure and data structure 1.}
\label{fig:neurotree1}
\end{figure*}

\subsection{How do we design the structure of the neurotree?}
\label{sec:neurotree_architecture}

Fig. \ref{fig:mlp-nt} shows a MLP and the corresponding neurotree. If we learn this neurotree structure through the level association networks (Section \ref{sec:gtlan}), the MLP model becomes a special case for level association networks (LANs). If the input exists only in the leaf node, it will pass through the same operation process as in the MLP.

Fig. \ref{fig:rnn-nt} shows an RNN and the corresponding neurotree.
A neurotree with only one child in succession is like a time-series dataset.
In addition, subtasks ($\Phi_{s}$) can be performed at each node, which allows tasks such as predicting the next input in RNNs. The RNNs will become a special case if we learn this neurotree with recursive association networks (RANs) as in Section \ref{sec:gtran}.

The GNN is a type of network that can be represented by a neurotree; it can train the relational information (Fig. \ref{fig:gnn-nt}).
The level of each neurotree has the same meaning as the number of layers of GNNs.
If we train this neurotree structure, it becomes a special case that corresponds to the last layer and readout process in the GNNs.

Finally, RecNNs can be used to express hierarchical information.
Existing tree datasets can be learned without major modifications using a neurotree, as shown in Fig. \ref{fig:recnn-nt}. The information delivery structure will be the same if we express tree datasets using an identity matrix ($\mathbf{I}$) as relationships between children ($\mathbf{A}_{c}$) and learn them using RANs.

\begin{figure}[H]
\centering
\begin{subfigure}{0.50\textwidth}
    \centering
    \includegraphics[height=2.8cm]{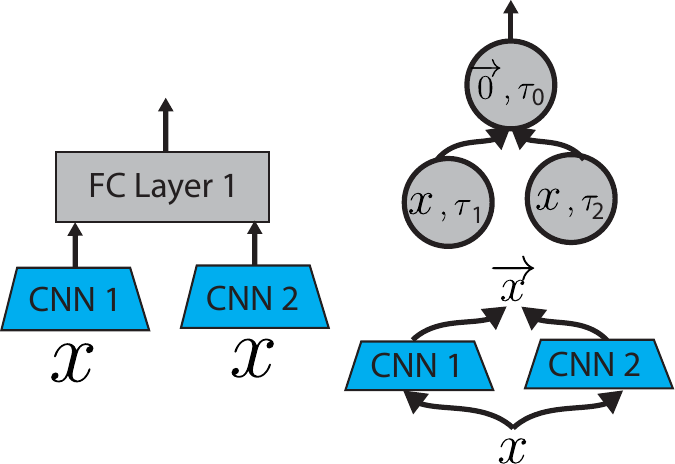}
    \caption{2-stream CNN and neurotree}
    \label{fig:2-stream-CNN}
    
\end{subfigure}
\hfill
\begin{subfigure}{0.40\textwidth}
    \centering
    \includegraphics[height=2.8cm]{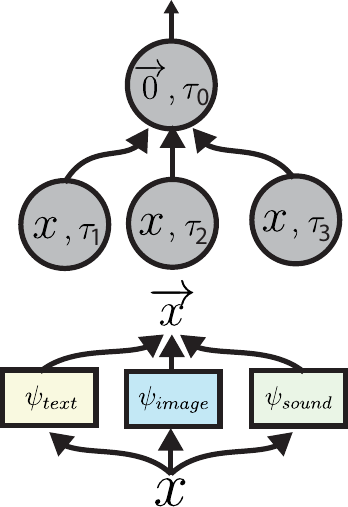}
    \caption{Connecting existing models (multimodal).}
    \label{fig:3-connecting}
\end{subfigure}
\caption{ Model structure and data structure 3.}
\label{fig:neurotree3}
\end{figure}

We can also use multiple feature extraction networks together.
For example, the same operation process will be performed if we use a convolutional neural networks (CNNs) for feature extraction ($\psi$) as shown in Fig. \ref{fig:cnn-nt} and replace the MLP process with an LAN.
In addition, it is possible to express the structure of an n-stream CNN.
Fig. \ref{fig:2-stream-CNN} illustrates models that each perform the functions of two CNN models and one concatenation process. The neurotree can express these models as tree data.
Most existing neural network models are fixed in a specific structure because they use fixed layers.
These fixed models can only perform within a specific domain or a specific task.
In contrast, various structures can be learned simultaneously using a neurotree structure.

This neural network structure can be applied by connecting various existing neural networks (e.g., VIT\cite{dosovitskiy2020image}, Word2Vec\cite{mikolov2013efficient}, and GPT\cite{brown2020language}) with multifeature extraction networks ($\Psi$) as shown in Fig. \ref{fig:3-connecting}.

Therefore, the existing basic neural network models (multilayer, recurrent, convolutional, recursive, and graph) can be represented in a neurotree, and it depends on how the tree structure is constructed.

Finally, Fig. \ref{fig:multi-parent-recursive-networks} shows the information delivery structure of a neurotree. Figure \ref{fig:multi-parent-gnn} shows the form of a fully connected neurotree for each level, and this structure is possible because it can have multiple parent nodes. 

A tree delivers information only to the parent node, as in Fig. \ref{fig:recnn-nt}. Therefore, the information cannot be delivered directly to a grandparent node. 
In contrast, the neurotree can have multiple parent nodes, maintain a feed-forward structure, and contain the overall propagation structure of the network model.

It is possible to directly deliver hidden states to distant parent nodes such as a grandparent or great-grandparent node; this constitutes a direct message delivery method between layers. Therefore, we determined the datasets that would need a structure that involves multiple parent nodes, and we selected appropriate source code datasets.

\subsection{Source Code Analysis and Representation of Neurotree}
We use relatively complex structured datasets to transform the structure into a neurotree and perform a simple experiment to demonstrate that information from more complex structures and more domains can be represented naturally in a single neural network and learned simultaneously. 

There is a wide variety of information in the source code,  e.g., CFGs, ASTs, and data flow dependencies\cite{PUHRWESTERHEIDE1979117}; however, it is difficult to learn all of this information simultaneously using neural networks. We introduce the process of parsing source code containing relatively complex structures and various information into CFGs and ASTs, and we transform them into neurotrees to represent more diverse information.

A representation of the process of learning dynamic information when executing code through static analysis is presented below.
In this process, information such as primary values, source code execution paths, memory load and store operations, function definitions and calls, and node-types of ASTs, are included.

\begin{figure}[h]
\centering
\begin{subfigure}{0.90\textwidth}
    \centering
    \includegraphics[height=4.2cm]{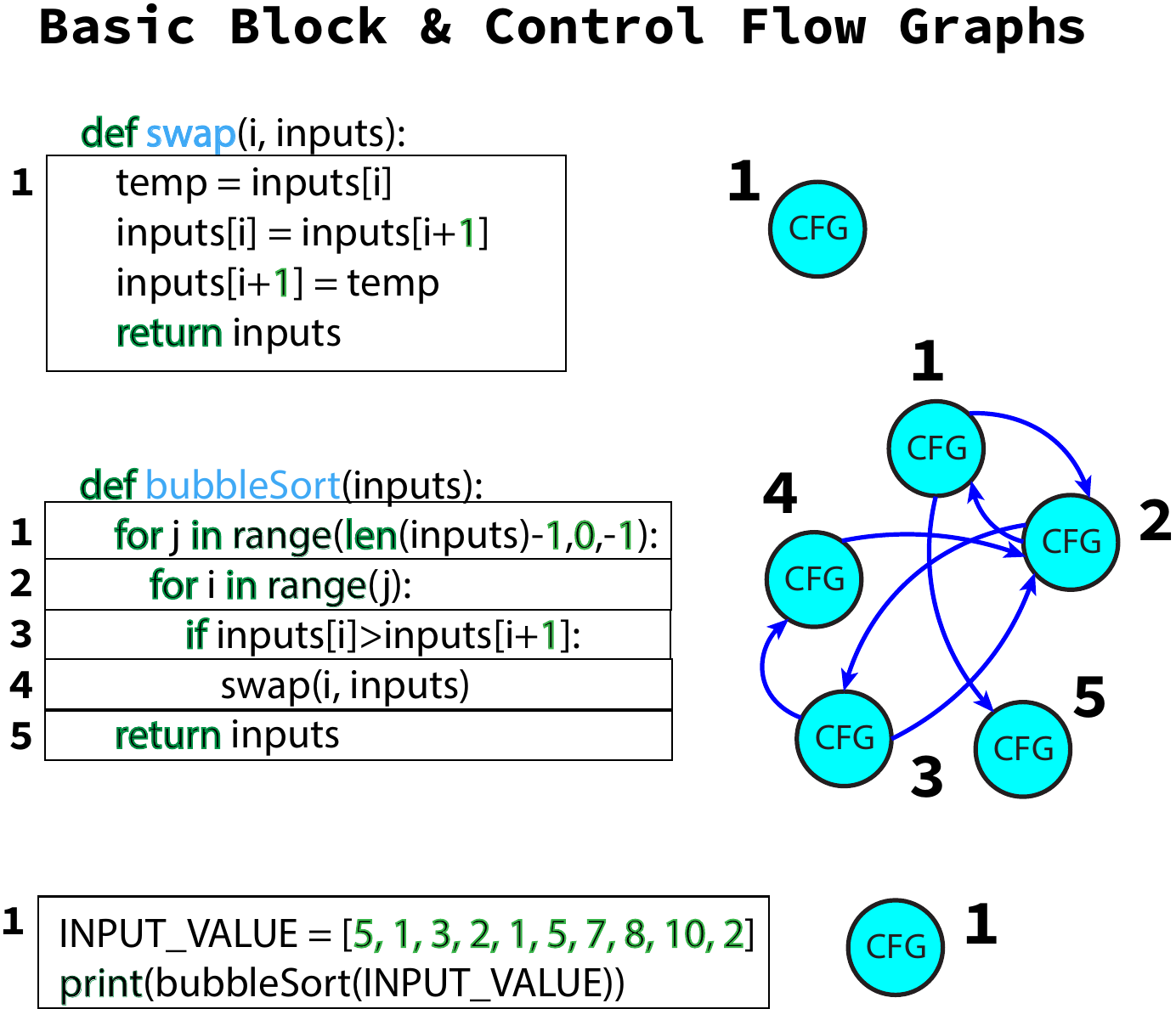}
    \caption{ Control flow graphs of functions }
    \label{fig:cfg_function}
\end{subfigure}
\hfill
\begin{subfigure}{0.90\textwidth}
    \centering
    \includegraphics[height=4.2cm]{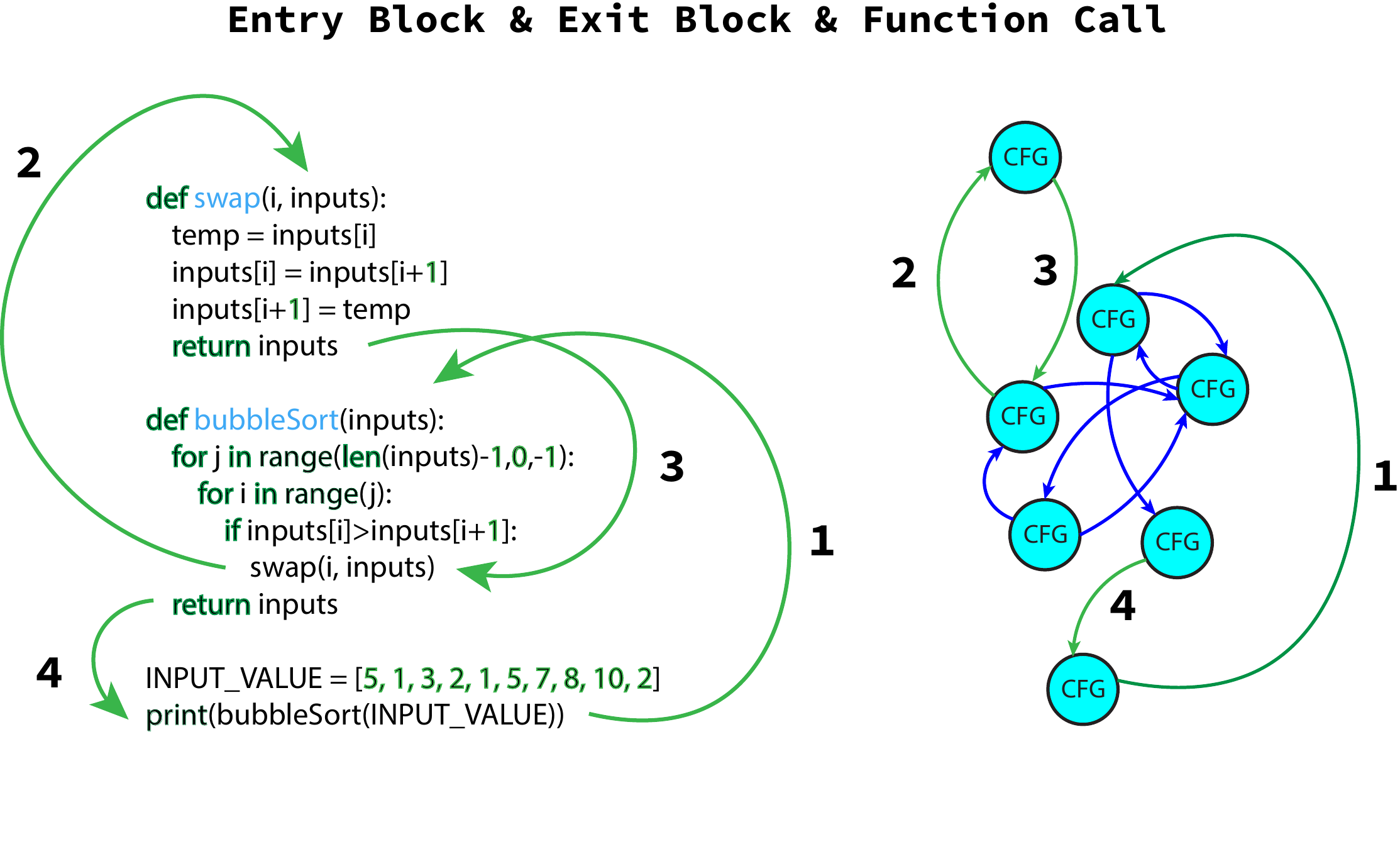}
    \caption{ Graph of all code execution paths }
    \label{fig:cfg_function_call}
\end{subfigure}
\caption{ Relationships ($\mathbf{A}_{c}$) between children (control-flow graphs and function call). }
\label{fig:controlflowgraph}
\end{figure}

The first step is to generate a CFG composed of basic blocks (Fig. \ref{fig:cfg_function}). We create a CFG with all functions. When a function is called, it is connected from the basic block of the code statement that called the function to the entry block of the CFG of the called function (Fig\ref{fig:cfg_function_call}). Further, a connection is made from the exit block of the called function to the basic block calling the function.

the function is connected from the basic block calling the function to the entry block of the function, and from the exit block of the function to the basic block calling the function.
Therefore, this CFG is used as the $\mathbf{A}_{c}$ of the root node; the basic blocks are used as child nodes ($\mathbf{C}$).
This structure is shown in Fig. \ref{fig:controlflowgraph}. There are code statements in this basic block parsed using an AST, and the following steps are performed.

\begin{algorithm}[h]
\caption{Case 1: Assign} 
\label{algo:assign_code}
\begin{algorithmic}
\State x = 3
\State y = x
\end{algorithmic}
\end{algorithm}

The code in Algorithm \ref{algo:assign_code} is used when an assign operation is performed and loaded.
An AST node for the assignment stores the variable name and value stored in the variable name as a child node.
We can store the value in the memory and load the stored value from the memory using the variable name when we execute the code.

\begin{figure}[ht]
\centering

\begin{subfigure}{0.90\textwidth}
    \centering
    \includegraphics[height=3.9cm]{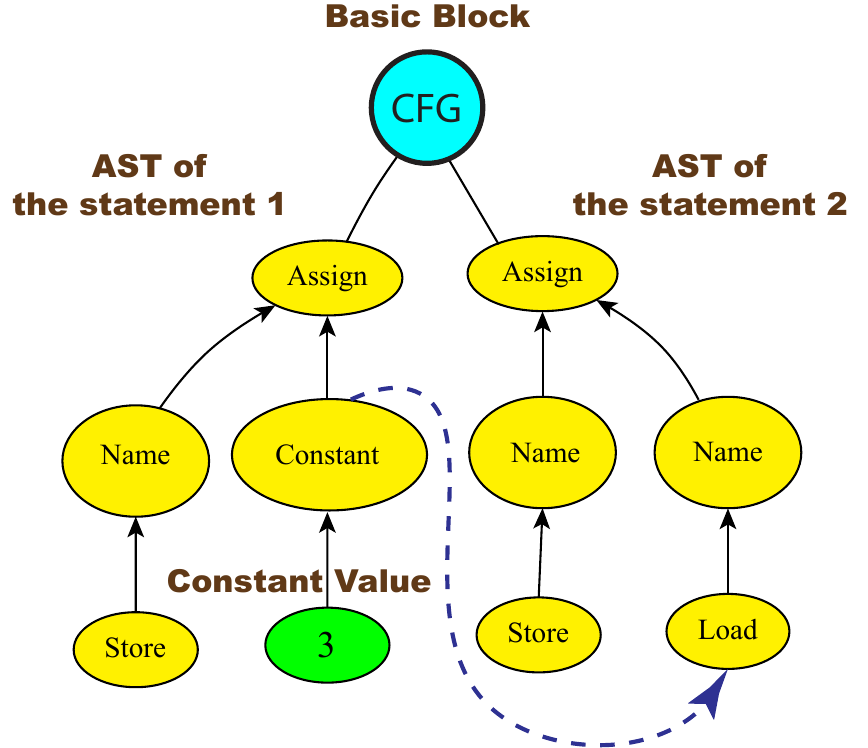}
    \label{fig:ast_assign_neurotree}
\end{subfigure}
\caption{ Nodes with multiple parents (assign). }
\label{fig:assign_neurotree}
\end{figure}
We store the assigned value in the dictionary with the variable name when an assignment occurs. The variable name is key, and the assigned value is the dictionary value.
In addition, the dictionary is called with the variable name to connect the called nodes like child nodes on nodes that load with the same variable name. 
Further, body regions (e.g., while, for, and if regions) and function definitions are removed from the AST because they exist in other basic blocks; this is
illustrated in Fig. \ref{fig:assign_neurotree}.

This neural data structure allows each node to learn various domain information and node types.
An AST contains not only node type information but also constant values.

Thus, we can define and learn multifeature extraction networks $\Phi = \{\psi_{ast}, \psi_{constant} \}$ using two feature extraction networks that handle domain information for node types and constant values.

In the dataset we used, there were 70 AST node types ($\psi_{node-type}$).
We used the one-hot encoding process and a fully connected embedding layer reduced to 25 dimensions because it is assumed that there will be about two or three similar types for each type. A hyperbolic tangent function is used as the activation function.

Constant values ($\psi_{constant}$) can be abnormally large; therefore, we calculated quantile values.
All constant values appearing in the learning dataset showed values of approximately $0\leq x \leq 10^{9}$.
The Q1 and Q3 of these values were 0, but we applied the Q3 as an integer 1 rather than 0 (Because these codes use a lot of integers); these Q1 and Q3 values were used as the criteria for the maximum and minimum values, respectively.
Thus, the values below 0 were replaced with -1, and the values above 1 were replaced with 2; the remaining values were kept as they were. We added $\psi_{constant}$ to learn the scale value as the input.

In this process, we represent the node types, relationships between functions as $\mathbf{A}_{c}$, relationships of memory as multiparent nodes, and multiple domains as multiple feature extraction networks. This neurotree data structure represents all types of existing data (image, sound, time-series, graph, tree, source code, etc.); finally, it performs convolution from the leaf node to the root node through recursive propagation.
We used the recursive propagation method introduced in Section \ref{sec:Depthfirst}.

We did not perform the pretraining process as in the TBCNN study; instead, we performed training together with a subtask that predicts the node type of the parent node.

\subsection{ Multi-Task Networks }
\label{sec:multi_task}
\begin{figure}[ht]
\centering
\begin{subfigure}{0.90\textwidth}
    \centering
    \includegraphics[height=5.7cm]{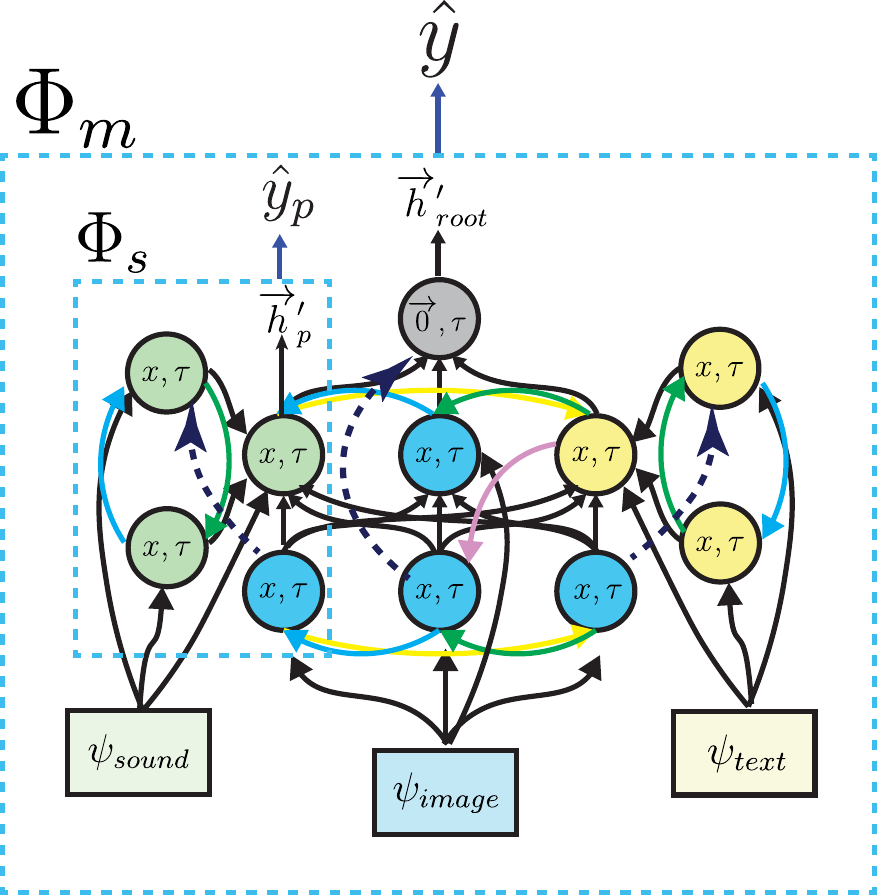}
\end{subfigure}
\caption{Subtask and main task.}
\label{fig:main-sub-task}
\end{figure}

There are two groups of tasks defined in this study: main tasks and subtasks. The main tasks refer to tasks performed only in the root neuronode, and the subtasks refer to simple tasks performed at each node level.

\begin{equation}
\hat{y}_{o} = \Phi_{s}(\overrightarrow{h}'_{o}, \tau_{so})
\label{eq:multi-sub-task}
\end{equation}

It seems appropriate to assign simple tasks as subtasks performed at the node level. TBCNNs have been effective in the subtask of predicting a parent node\cite{mou2014tbcnn}.
From a time-series data perspective, these tasks are equivalent to tasks that predict the input of the next step, and therefore, they are used in RNNs (Fig \ref{fig:neurotree1}(b)).
Thus, a function that performs node-level tasks by connecting to other networks based on the subtask is defined as $\Phi_{s}$.

\begin{equation}
\hat{y} = \Phi_{m}(\overrightarrow{h}'_{root}, \mathbf{NT}', \tau_{m})
\label{eq:multi-main-task}
\end{equation}
In this approach, the task of the root node is special; the tasks performed in the root node can be combined with other neural networks such as deep Q-networks(DQNs)\cite{mnih2013playing} to perform actions or relate to various thinking activities.
These tasks are defined as main tasks; therefore, in this study, a function that coordinates with other neural networks based on the main task is defined as $\Phi_{m}$.

This $\Phi_{m}$ performs the task by inputting the hidden state of the root node ($\overrightarrow{h}'_{root}$) and overall neurotree ($\mathbf{NT}'$) resulting from DFC (Section \ref{sec:Depthfirst}).

The information that can be used inside the neurotree can differ for different main tasks; the classifier uses the hidden state in the root node when performing the classification task. 
The overall neuronode information needs to be entered when performing a restoration task such as using a decoder(e.g., an autoencoder\cite{hinton2006reducing}).

The main task in this paper comprises the classification in various domains.
Other main tasks in progress will be introduced in future studies~\cite{kim2021deductive, kim2021memory, kim2021imagine}.

In AANs, $\tau_{d}$ indicates multidomain information, building a neurotree ($\mathbf{NT}$) indicates multimodal learning, and $\tau_{s}, \tau_{m}$ denotes multi-tasks.

\subsection{ Mini-Batch Propagation Algorithms: DFC and DFD }
\label{sec:Depthfirst}
\begin{figure}[ht]
\centering
\begin{subfigure}{0.47\textwidth}
    \centering
    \includegraphics[height=2.8cm]{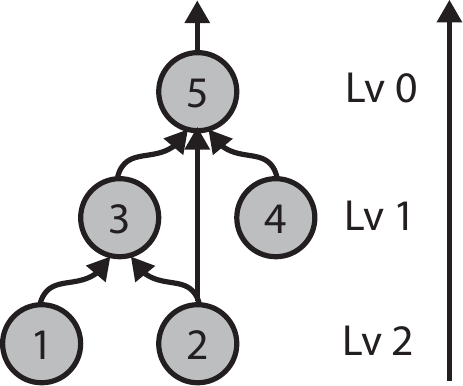}
    \caption{Post-order and left-first}
    \label{fig:dfc}
\end{subfigure}
\hfill
\begin{subfigure}{0.47\textwidth}
    \centering
    \includegraphics[height=2.8cm]{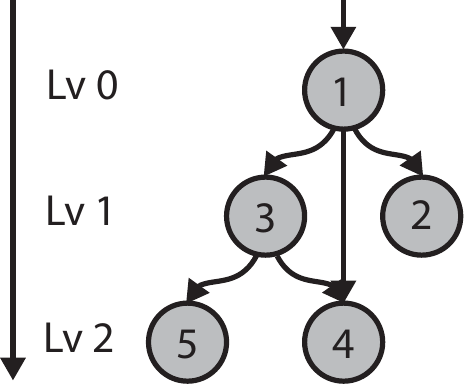}
    \caption{Pre-order and right-first}
    \label{fig:dfd}
\end{subfigure}
\caption{DFC and DFD}
\label{fig:dfcdfd}
\end{figure}

DFC is a recursive propagation methodology that propagates from the nodes of the deepest level of the neurotree to the root node using the DFS algorithm.
In a neurotree, one neuronode can have multiple parents, and therefore, this algorithm can visit one neuronode several times.
Thus, the convolution is performed only once in a neuronode, and upon revisiting, the previously calculated hidden state information is delivered to the parent node.

There is an additional point to consider when performing \textbf{batch learning} for a neurotree with multiple parent structures.
We deliver the hidden state information performed in the neuronode such that \textbf{stores} it in the neuronode to simplify the operation.
We call the information stored in that node because it is a reference object when we need this calculated information. The DFS algorithm makes it possible to propagate to multiple parent nodes in a batch neurotree structure. 

Unlike in RecNNs, which only deal with existing tree structures, it is possible to learn relationships between multiple sets of sibling nodes that are learned simultaneously. 
The features for various domains are learned together. 

This study shows that the structures of existing neural networks can be expressed in data (neurotree) and learned.
Therefore, we describe depth-first deconvolution (DFD), a propagation method that traverses in the reverse order, to show that autoencoder or bidirectional models are feasible. We describe these algorithms in detail in Appendix \ref{appendix:dfc-algorithm}, \ref{appendix:dfd}.

\begin{algorithm}[ht]
\caption{Propagation (for autoencoder models).}
\label{algorithm:propagate}
\begin{algorithmic}[1]
            \Function{propagate}{$\mathbf{BNT}$}
                \State {$\overrightarrow{\mathbf{h}}_{root},\mathbf{BNT}'$ $\gets$ {\Call{DFC}{$\mathbf{BNT}.\mathbf{NN}_{root},0$}}} 
                \State $\hat{\mathbf{y}}$ $\gets$ $\Phi_{m}[\tau_{m}]${($ \protect\overrightarrow{\mathbf{h}}_{root},\protect\mathbf{BNT'}.\protect\mathbf{NN}_{root}$)} 
            \State \Return $\hat{\mathbf{y}}, \overrightarrow{\mathbf{h}}_{root},\mathbf{BNT}'$
            \EndFunction
            \State
            \Function{$\phi_{autoencoder}$}{$\protect\overrightarrow{\mathbf{h}}_{root}, \mathbf{NN}_{root}$} 
                \State $\protect\overrightarrow{\mathbf{x}}_{root} \gets \protect\mathbf{NN}_{root}.\overrightarrow{\mathbf{x}}$
                \State \Return \Call{DFD}{$\protect\overrightarrow{\mathbf{x}}_{root}, \protect\overrightarrow{\mathbf{h}}_{root},\protect\mathbf{BNT'}.\protect\mathbf{NN}_{root},0$} 
                
            \EndFunction
\end{algorithmic}
\end{algorithm}
$\mathbf{BNT}$ refers to a batch neurotree. By defining the DFC and DFD as the propagation algorithm (Algorithm \ref{algorithm:propagate}), it becomes possible to freely encode or decode the neurotree.
We implemented the DFD algorithm, but not in this work.
In the next study, we introduce its usage in detail\cite{kim2021memory}.
We show the relationship between DFC and DFD, and the possibility of expanding them to the widely used autoencoder\cite{hinton2006reducing} models. 
Further, the root node performs an additional main task, which unlike a subtask, is performed only in the root node.

The main task can combine various models such as a decoder, generator, and classifier. 
In this study, our experiments focused on classification as the main task, and we leave other contents as future work\cite{kim2021deductive, kim2021memory, kim2021imagine}.

\subsubsection{Multi-Batch Feature Extraction Method}
\label{sec:multi-mini-feature-extraction}
\begin{figure}[ht!]
    \centering
    \includegraphics[height=3.7cm]{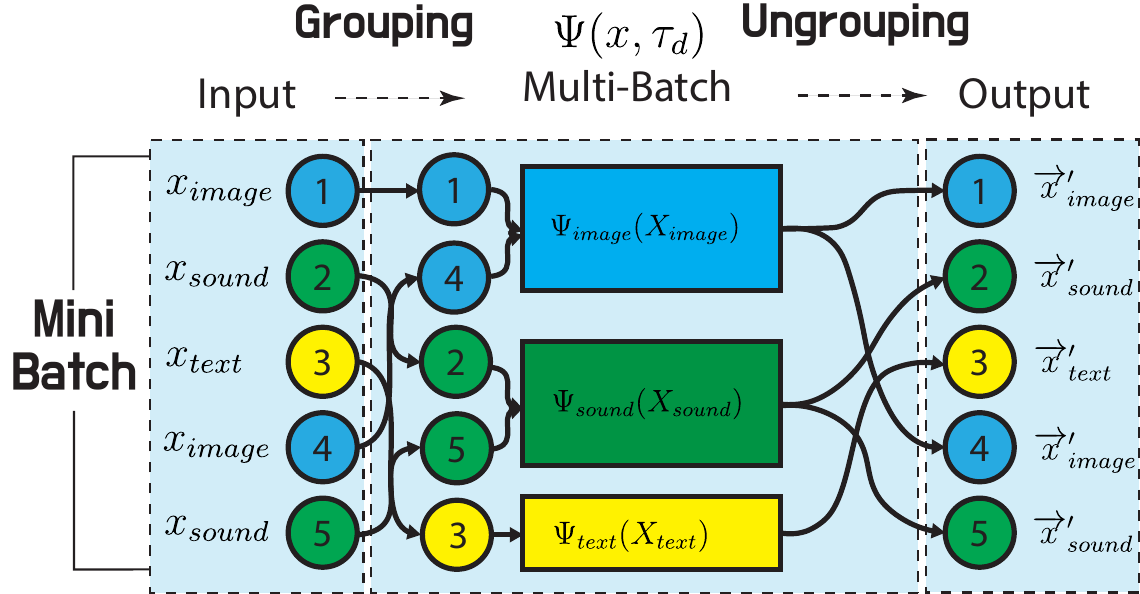}
    \hfil
    \caption{ Multifeature extraction networks for multi-mini-batch training. }
\label{fig:febatch}
\end{figure}
This algorithm extracts features from batch neuronodes during batch learning.
Several types of domains exist within batch neuronodes, and the feature extraction networks applied to each domain are different; therefore, it is very inefficient to extract the sample features one by one.
Thus, we perform feature extraction by grouping based on domain and ungrouping to find the original batch index (as Fig. \ref{fig:febatch}).

This method divides mini-batch samples by domain, and the domain-mini-batch size varies each time (Fig. \ref{fig:febatch}). Thus, we recommend using weight standardization\cite{qiao2019micro} or group normalization\cite{wu2018group} to avoid causing batch size to affect batch normalization\cite{ioffe2015batch}. We describe these algorithms in detail in Appendix \ref{algorithm:batch_type_embedding}.

Multi-tasks ($\Phi_{s}, \Phi_{m}$) are implemented to perform mini-batch learning with algorithms similar to these processes.
\subsection{ Artificial Association Network Models: RAN and LAN Series }
\label{sec:network}
The main difference between the two types of AANs (RANs and LANs) is that the former uses the same weight parameters at all levels, whereas the latter use different weight parameters at each level.
The difference caused by this structure is provided in Section \ref{appendix:compare}.

Further, we introduce \textbf{the convolution performed in the neuronode of the o-th order during the DFC process} of propagation in the neurotree.
These networks can be used as if they were association cells or layers.
\subsubsection{ Association Cell: Recursive Association Networks }
\label{sec:gtran}
We describe the process of the RAN cell learning the neurotree.
Recursive association models can be expressed as $\mathbf{RAN}=\{\mathbf{W}, \Psi, g, \sigma \}$ $\mathbf{W}\in\mathbb{R}^{F' \times (F+B+F')}$.
Here, $F$, $B$, $F'$, $\Psi$, \(g\), and $\sigma$ denote the feature size of the node, domain bias size, hidden size of the hidden state received from the children, multifeature extraction process (as in (\ref{eq:feature_extraction})), aggregate function, and activation function, respectively.

This DFC (Section \ref{sec:Depthfirst}) process propagates from the leaf nodes to the root nodes, and it has a feed-forward structure that delivers from the deepest level to the top level.

$o$, $i$, and $N_{o}$ represent the propagation order of DFC, order of the child node, and number of the children of $\mathbf{NN}_{o}$, respectively. Therefore, $N_{o} = |\mathbf{C}_{o}|$. $x_{o}$ represents the node input of $\mathbf{NN}_{o}$.
\begin{equation}
\label{eq:gtran1}
\overrightarrow{x}'_{o} = [\psi_{\tau_{do}}(x_{o}),onehot(\tau_{do})] = \Psi({x}_{o},\tau_{do}), \overrightarrow{x}'_{o} \in \mathbb{R}^{F+B}
\end{equation}
First, $\Psi$ extracts a feature vector ($\overrightarrow{x}'_{o}$) considering the domain information from $x_{o}$. 
Then, the domain bias is concatenated with $\overrightarrow{x}_{o}$(Section \ref{sec:multi_feature_extraction}).
\begin{equation}
\label{eq:gtran2}
\overrightarrow{h}_{oi} = \mathbf{NN}_{o}.\mathbf{C}_{i}.\overrightarrow{h}'
\end{equation}
\begin{equation}
\label{eq:gtran3}
\mathbf{h}_{o} = \mathbin\Vert_{i=0}^{N_{o}}\overrightarrow{h}_{oi}
\end{equation}
The o-th neuronode acquires the hidden state information from all children.
The structure used to receive the hidden state from the i-th child can be expressed as \ref{eq:gtran2}.
This process is repeated $N_{o}$ times to obtain the hidden state from all the children, and then, $\mathbf{h}_{o}$ is created in the form of a stack ($\mathbf{h}_{o}\in\mathbb{R}^{N_{o} \times F'}$).
Thus, $\mathbf{h}_{o}$ contains all information about the children.
\begin{equation}
\label{eq:gtran4}
\overrightarrow{h}_{o} = g(\tilde{\mathbf{D}}_{co}^{-\frac{1}{2}}\tilde{\mathbf{A}}_{co}\tilde{\mathbf{D}}_{co}^{-\frac{1}{2}}(\mathbf{h}_{o}))
\end{equation}
Now, all information received from the children is aggregated ($g$).
In this aggregation process, the relationships between the children and received hidden states of the children are considered together.
Thus, we stored the relationships between children as $\mathbf{A}_{c}$ in the neuronode.
An identity matrix ($\mathbf{I}$) is used if there are no relationships between children.

Thus, the children's hidden states ($\mathbf{h}_{o}$) and relationship graphs ($\mathbf{A}_{c}$) perform graph convolution and readout to aggregate their information; the aggregated information is expressed as $\overrightarrow{h}_{o}$.

We applied the GCN methodology\cite{kipf2016semi}, which is useful in the GNN field.
Therefore, we expressed the relational term as $\tilde{\mathbf{A}}_{co}=\mathbf{A}_{co} + \mathbf{I}$, where $\mathbf{I}$ represents the identity matrix.

If we express the connection value as 1, only the more connected nodes have scale values larger than those of the other nodes.
Therefore, $\tilde{\mathbf{D}}_{co}^{-\frac{1}{2}}\tilde{\mathbf{A}}_{co}\tilde{\mathbf{D}}_{co}^{-\frac{1}{2}}$ is applied using the order matrix $\tilde{\mathbf{D}}_{co}$ of $\tilde{\mathbf{A}}_{co}$ as a method of normalizing the relationship matrix in (\ref{eq:gtran4}). 
\begin{equation}
\label{eq:gtran5}
\overrightarrow{h}'_{o} = \sigma([\overrightarrow{x}'_{o},\overrightarrow{h}_{o}]\mathbf{W}^{T})
\end{equation}
Finally, we perform convolution using $\overrightarrow{x}'_{o}$ and $\overrightarrow{h}_{o}$.
The hidden state ($\overrightarrow{h}_{o}$) is concatenated with $\overrightarrow{x}'_{o}$ to obtain $[\overrightarrow{x}'_{o}, \overrightarrow{h}_{o}] \in \mathbb{R}^{F+B+F'}$ through the concatenation process as $[,]$.
The children's aggregated hidden state ($\overrightarrow{h}_{o}$) and the ($\overrightarrow{x}'_{o}$) of the current node perform convolution to obtain the current node's hidden state ($\overrightarrow{h'}_{o}$).
The convolution form is similar to that of RNNs.
If there is no input or no children, $\overrightarrow{0}_{o}$ is used instead of $\overrightarrow{x}'_{o}, \overrightarrow{h}_{o}$.
\begin{equation}
\label{eq:gtran6}
\overrightarrow{h'}_{o} = \sigma([\overrightarrow{x}'_{o}, g((\tilde{\mathbf{D}}_{co}^{-\frac{1}{2}}\tilde{\mathbf{A}}_{co}\tilde{\mathbf{D}}_{co}^{-\frac{1}{2}})\mathbf{h}_{o})]\mathbf{W}^{T})
\end{equation}
Finally, we express the processes (\ref{eq:gtran1}, \ref{eq:gtran2}, \ref{eq:gtran3}, \ref{eq:gtran4}, and \ref{eq:gtran5}) in (\ref{eq:gtran6}).
We set $F$ as 128, $F'$ as 128, and $\sigma$ as ReLU\cite{nair2010rectified}; g was the maxpool. The size of $B$ varied depending on the experiment.

If no sibling nodes exist for any nodes, (\ref{eq:gtran4}) may be omitted, and the RAN is mathematically identical to an RecNN.

\subsubsection{Association Layers: Level Association Networks}
\label{sec:gtlan}
The LAN structure is remarkably similar to that of an RAN. An LAN is a type of AAN that uses different weight parameters by level to make the FC layer and MLP structure a special case. The LANs are composed as follows: $\mathbf{LAN}=\{\{\mathbf{W}_{0},...\mathbf{W}_{LV-1}\},\Psi, g, \sigma\}$.
\begin{equation}
\label{eq:gtlan}
\overrightarrow{h'}_{o} = \sigma([\overrightarrow{x}'_{o}, g((\tilde{\mathbf{D}}_{co}^{-\frac{1}{2}}\tilde{\mathbf{A}}_{co}\tilde{\mathbf{D}}_{co}^{-\frac{1}{2}})\mathbf{h}_{o})]\mathbf{W}^{T}_{lv})
\end{equation}
The mathematical expression (\ref{eq:gtlan}) of an LAN is like (\ref{eq:gtran6}).
The difference between LANs and RANs is that these weight parameters are used as $\mathbf{W}_{lv}$ in the process of (\ref{eq:gtran5}); thus, the weight parameters are used differently for each level.

The weight parameters $\mathbf{W}_{lv}\in\mathbb{R}^{F'_{lv} \times (F_{lv}+F'_{lv+1})}$ control each level; the network in charge of each level is called the level layer.
The input size of the level is $F_{lv}$, its output size is $F'_{lv}$, and the output size of the child is $F'_{lv+1}$.

Thus, it is possible to adjust the input, i.e., the hidden size; this type of model can learn depth-limited trees, where $LV$ denotes the maximum depth. We set $F_{lv}$ and $F'_{lv}$ to 128 for all LVs like that in the RAN in the experiment.

\begin{equation}
\label{eq:spcase4}
Multi\ layer = fc_{N}(..fc_{2}(fc_{1}(\overrightarrow{x}))) = \mathbf{LAN}.DFC(\mathbf{NN}_{0})
\end{equation}
The LAN is mathematically identical to the FC layer and MLP if the input size of $F_{lv}$ is 0 and there are no sibling nodes of the current node. $\mathbf{I}$ represents the identity matrix and $DFC$ denotes the DFC.
The depth of the tree indicates the number of layers of the MLP. Thus, the MLP can be considered as a special case of LAN.
\subsection{Architecture Comparison: LAN and RAN }
\label{appendix:compare}
\begin{figure}[h]
\begin{floatrow}
\capbtabbox{%
\footnotesize
\begin{tabular}{l
              c
              @{\hspace*{2.0mm}}c
              @{\hspace*{2.0mm}}c
              @{\hspace*{2.0mm}}c
              @{\hspace*{2.0mm}}c
              }
\toprule
    Name     & LAN & RAN \\
    \midrule
    Type of layers  & Level layer & Cell \\
     Number of W parameters & Number of levels & 1 \\
    Depth-limited & Fixed & Not fixed   \\
    Number of parameters & Higher &  Lower \\
    Input size by level & Adjustable & Fixed \\
    Special cases & FCNN, MLP & RNN, RecNN \\
    Aggregation process & \multicolumn{2}{c}{Readout with relation term as GNN}\\
\bottomrule
\end{tabular}
}{%
 \caption{Architecture comparison.}
\label{table:Comparison}
}
\end{floatrow}
\end{figure}
In the case of RANs, fewer parameters are used than that in LANs because the association cells used for all neural nodes are the same.

Further, the size of the information is not flexible because the output size of the feature vectors of multifeature extraction networks and size of the hidden state are the same at all levels.

By contrast, in the case of LANs, it is possible to use a different feature vector size because $\mathbf{W}_{lv}$ is different for each level, and because the concept of the ``level layer'' is used.
However, there is a maximum depth of the neurotree that can be processed because the weight parameters must be defined for all levels.

We believe that it is appropriate to use LANs when certain domain information is associated with each level; RANs when various domains are involved in all processes.

In addition, when we designed $\mathbf{AANs}$, $\mathbf{LANs}$ were designed to represent FC layer, multilayer structures as trees, and $\mathbf{RANs}$ were designed to represent \ recurrent, recursive structures as trees (Section \ref{sec:neurotree_architecture}, \ref{sec:network}). The results are compared in Table \ref{table:Comparison}.

\subsection{ Attention Models }
\label{sec:attention_models}
GATs\cite{velivckovic2017graph}, a type of model that has been useful recently, were integrated to mimic the attention process in the human brain.
\subsubsection{ Recursive Attentional Association Networks }
\label{subsec:RAANn}
We introduce recursive attentional association networks (RAANs) that learn importance through attention by slightly modifying the expression of the RAN.
An RAAN is composed of $\mathbf{RAANs}=\{\mathbf{W}, \Psi,g, \sigma, \sigma_{a}, \overrightarrow{a} \}$, that is, $\{\sigma_{a}, \overrightarrow{a}\}$ were added to the RAN structure. A parameter for the attention mechanism is added ($\mathbb{R}^{2F'} \times \mathbb{R}^{2F'} \rightarrow \mathbb{R}$) and LeakyReLU\cite{xu2015empirical} was used as the attention activation function in the same way as in GATs.
$\mathcal{N}_{pq}$ represents a set of nodes connected to the i-th child's node in the $A_{co}$ of $\mathbf{NN}_{o}$, and we can express this as 
\begin{equation}
\label{eq:attention_matrix}
\alpha_{oij} = \frac{\mathbf{exp}(LeakyReLU(\overrightarrow{\mathbf{a}}^{T}[\overrightarrow{h}_{oi},\overrightarrow{h}_{oj}]))}{\sum_{k\in \mathcal{N}_{oi}}\mathbf{exp}(LeakyReLU(\overrightarrow{\mathbf{a}}^{T}[\overrightarrow{h}_{oj},\overrightarrow{h}_{ok}]))}
\end{equation}
With the attention methodology introduced in GATs\cite{velivckovic2017graph}, RAANs learn how the j-th node is important to the i-th node.
This information is replaced with the part to which the adjacency matrix is connected.

Therefore, the RAAN is given as
\begin{equation}
\label{eqn:attention1}
\overrightarrow{h'}_{o} = \sigma([\overrightarrow{x}'_{o}, g((\mathbf{A}_{co}\odot\mathbf{\alpha}_{o})\mathbf{h}_{o})]\mathbf{W}^{T})
\end{equation}
In (\ref{eqn:attention1}), $\odot$ indicates the hadamard product, and this process is similar to that reported in (\ref{eq:gtran6}).

This process becomes a cell and delivers the result from the leaf node to the root node.
\subsubsection{Level Attentional Association Networks}
The level attentional association network (LAAN) model was modified based on the  LAN, and the GAT attention mechanism was applied.
This model can be expressed as $\mathbf{LAAN}=\{\{\mathbf{W}_{0},...\mathbf{W}_{LV-1}\},\{\overrightarrow{a}_{0},...\overrightarrow{a}_{LV-1}\},\Psi, g, \sigma, \sigma_{a}\}$, and $\{\sigma_{a}, \overrightarrow{a}_{lv}\}$ are added to the LAN ($\overrightarrow{a_{lv}}\in\mathbb{R}^{2F^{'}_{lv}}$).
Unlike RAANs, the LAANs can design varied sizes of $F, F'$ to match the $lv$.
\begin{equation}
\label{eq:level_attention_matrix}
\alpha_{oij} = \frac{\mathbf{exp}(LeakyReLU(\overrightarrow{\mathbf{a}}^{T}_{lv}[\overrightarrow{h}_{oi},\overrightarrow{h}_{oj}]))}{\sum_{k\in \mathcal{N}_{oi}}\mathbf{exp}(LeakyReLU(\overrightarrow{\mathbf{a}}^{T}_{lv}[\overrightarrow{h}_{oi},\overrightarrow{h}_{ok}]))}
\end{equation}
\begin{equation}
\label{eqn:levelattention1}
\overrightarrow{h'}_{o} = \sigma([\overrightarrow{x}'_{o}, g((\mathbf{A}_{co}\odot\mathbf{\alpha}_{o})\mathbf{h}_{o})]\mathbf{W}^{T}_{lv})
\end{equation}
This model is designed to select critical features; it is similar to the RAANs discussed in Section \ref{subsec:RAANn}.
\subsection{ Gated Models }
\label{sec:gatedmodel}
The longer the depth of time, the more difficult it is for the RNN to propagate the error. Thus, we add the gated recurrent unit (GRU\cite{chung2014empirical}) model.
We designed the gated association unit (GAU) by slightly modifying the RAN.
\subsubsection{ Gated Association Unit }
The GAU is a model that changes the process of (\ref{eq:gtran5}) used in RANs to the GRU model.\label{subsec:GAU}
\begin{equation}
\label{eqn:GGTAN}
\overrightarrow{h}_{o} = g((\tilde{\mathbf{D}}_{co}^{-\frac{1}{2}}\tilde{\mathbf{A}}_{co}\tilde{\mathbf{D}}_{co}^{-\frac{1}{2}})\mathbf{h}_{o})
\end{equation}
This process is the same as aggregating a child’s hidden state in (\ref{eq:gtran4}).
\begin{equation}
\label{eqn:gatedgcn1}
\overrightarrow{r}_{o} = \sigma(\mathbf{W}_{j}\overrightarrow{h}_{o} + \mathbf{U}_{j}\overrightarrow{x}'_{o})
\end{equation}
\begin{equation}
\label{eqn:gatedgcn2}
\overrightarrow{u}_{o} = \sigma(\mathbf{W}_{u}\overrightarrow{h}_{o} + \mathbf{U}_{u}\overrightarrow{x}'_{o})
\end{equation}
\begin{equation}
\label{eqn:gatedgcn3}
\tilde{h}_{o} = tanh(\mathbf{W}(\overrightarrow{h}_{o} \odot \overrightarrow{r}_{o}) + \mathbf{U}\overrightarrow{x}'_{o})
\end{equation}
\begin{equation}
\label{eqn:GAU}
\overrightarrow{h'}_{o} = (1-\overrightarrow{u}_{o}) \odot \overrightarrow{h}_{o} + \overrightarrow{u}_{o} \odot \tilde{h}_{o}
\end{equation}
Furthermore, the GRU learns using the input information and aggregated hidden state. 
\subsubsection{ Gated Attentional Association Unit }
\label{subsec:GAAU}
\begin{equation}
\label{eqn:GTAAN}
\overrightarrow{h}_{o} = g((\mathbf{A}_{co}\odot\mathbf{\alpha}_{o})\mathbf{h}_{o})
\end{equation}

\begin{equation}
\label{eqn:GAAU}
\overrightarrow{h'}_{o} = (1-\overrightarrow{u}_{o}) \odot \overrightarrow{h}_{o} + \overrightarrow{u}_{o} \odot \tilde{h}_{o}
\end{equation}
Finally, we introduce the GAAU model that combines the RAN model with the attention mechanism applied in Section \ref{sec:attention_models} and the gated model applied in Section \ref{sec:gatedmodel}.
This model combines the attention matrix and GRU to improve learning in time-series data using a deep neurotree.
We implemented a neural network model with a basic structure using the proposed process. Further, it is possible to apply the structures of models by slightly modifying the proposed structure. It is believed that a better performance can be achieved in this manner.

\subsection{End-to-End Multi-Deep learning}
\begin{figure}[h]
    \centering
    \includegraphics[height=2.8cm]{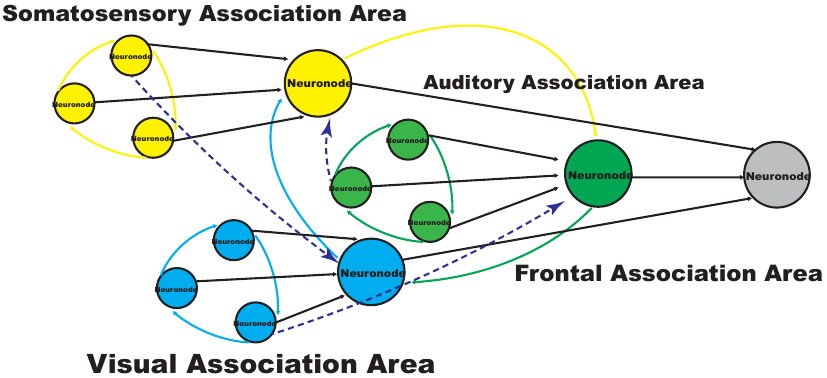}
    \hfil
    \caption{Neurotree association theory.}
\label{fig:artificial-association-networks-neurotree}
\end{figure}

\begin{equation}
\hat{y}_{o} = \Phi_{m}(\mathbf{AAN}.DFC(buildingNT(X)),\tau_{m})
\label{eq:end-to-end}
\end{equation}
To express this process, function ($f(x) \rightarrow y$) is created as an end-to-end structure. $buildingNT(X)$ was introduced in Section \ref{sec:neuro_datastructure}, \ref{sec:neurotree_architecture}, $DFC$ was introduced in Section \ref{sec:Depthfirst}, $\mathbf{AAN}$ was introduced in Section \ref{sec:multi_feature_extraction}, \ref{sec:network}, and $\Phi_{m}$ was introduced in Section \ref{sec:multi_task}.
\begin{equation}
\label{eq:world2infomation10}
sub\_task\_loss = \frac{\sum_{o}error(\mathbf{NN}_{o}.y, \mathbf{NN'}_{o}.\hat{y})}{sub\_task\_count}
\end{equation}
We calculate the loss of tasks to train the network. The target is compared with the outputs of the subtask stored in the neuronodes inside the neurotree to calculate the loss suitable for the subtask.
\begin{equation}
\label{eq:world2infomation11}
main\_task\_loss = error(y, \hat{y})
\end{equation}
Furthermore, the network is trained through back-propagation by calculating the loss for the main task.
\begin{equation}
\label{eq:world2infomation12}
loss = main\_task\_loss + sub\_task\_loss
\end{equation}
The artificial association models are as follows: the feature extraction models create the information from data that corresponds to the smallest unit of information in $\mathbf{NT}$.
The parent $\mathbf{NN}$ receives information from its children and performs the convolution operation with the relation term; the convolution output is a larger unit of information.
Finally, the root $\mathbf{NN}_{root}$ has all $\mathbf{NT}$ information, and it perform the convolution operation; the output is the largest unit of information as shown in Fig. \ref{fig:artificial-association-networks-neurotree}.

\textbf{ Supervised learning \;\;}
Supervised learning uses datasets with labels.
In this network, $\overrightarrow{h}_{root}$ was mapped to the dimension of the class using a FC layer ($F'$ to class count), and we calculated the log SoftMax and negative log likelihood loss for supervised learning as 

\begin{gather}
\hat{y}_{o} = log\_softmax(\Phi_{s}(\overrightarrow{h}_{o}',\tau_{s}))\\[5pt]
\hat{y} = log\_softmax(\Phi_{m}(\overrightarrow{h}_{root}',\tau_{m}))\\[5pt]
loss = nll\_loss(y, \hat{y}) + \frac{\sum_{o}nll\_loss(y_{o}, \hat{y}_{o})}{sub\_task\_count}
\end{gather}

We applied these processes to the classification of main task ($\Phi_{m}$) and prediction parent node for subtask ($\Phi_{s}$).
\section{Experiments and Results}
\label{sec:experimental}
\begin{figure}[h]
\begin{floatrow}
\capbtabbox{%
\footnotesize
\begin{tabular}{l
              c
              @{\hspace*{2.0mm}}c
              @{\hspace*{2.0mm}}c
              @{\hspace*{2.0mm}}c
              @{\hspace*{2.0mm}}c
              }
\toprule
\footnotesize Model & In & Out  & $\overrightarrow{a}$ & Norm & $\sigma$ \\ 
\midrule 
RAN    & 128  & 128 & - &  layer-norm & leaky-relu \\    
RAAN & 128 & 128 & (256,1) &  layer-norm  & leaky-relu \\
LAN    & 128 & 128 & - &  layer-norm & leaky-relu   \\     
LAAN & 128 & 128 & (256,1) & layer-norm  & leaky-relu    \\
GAU    & 128 & 128 & - & layer-norm & tanh  \\     
GAAU & 128 & 128 & (256,1) & layer-norm  & tanh \\
\midrule
Final & - & 128 & - & - & - \\
\bottomrule
\end{tabular}
}{%
  \caption{$\mathbf{AAN}$ parameters.}%
\label{table:artificial-association-networks-experiment-models}
}

\end{floatrow}
\end{figure}
\begin{figure}[ht]
\begin{floatrow}
\capbtabbox{%
\footnotesize
\begin{tabular}{l
              c
              @{\hspace*{2.0mm}}c
              @{\hspace*{2.0mm}}c
              @{\hspace*{2.0mm}}c
              @{\hspace*{2.0mm}}c
              }
\toprule

\scriptsize Name & Train & Test & Valid  & Classes & Structure  \\ 
\midrule 
MNIST & 50,000 & 10,000 & 10,000 & 10 & image    \\    
\scriptsize{SC} & 84,843 & 11,005 & 9,981 & 35 & sound  \\
IMDB    & 17,500 & 25,000 & 7,500 & 2 & text   \\     
SST & 8,544 & 2,210 & 1,101 & 2 & tree   \\
Algorithms  & 136 & 43 & 34 & 6 & tree, neurotree \\     
UPFD & 1,092 & 3,826 & 546 & 2 & graph \\
Iris & 96 & 30 & 24 & 3 & tabular \\

\bottomrule
\end{tabular}
}{%
  \caption{Datasets.}%
\label{table:experimental-datasets}
}
\end{floatrow}
\end{figure}
\begin{figure*}[t!]
    \centering
    \includegraphics[height=3.2cm]{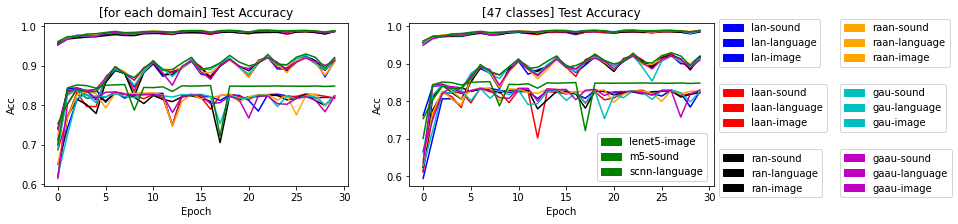}
    \hfil
    \caption{ Test accuracy of simultaneous learning (Experiment 1).  }
\label{fig:experimental-1-neurotree}
\end{figure*}

\textbf{Our goal is to express the information delivery process of existing models as a neurotree.} 
In all experiments, we set the seed number to 1234. The learning rate was $10^{-3}$ using the Adam optimizer\cite{kingma2014adam}, the scheduler was cosine annealing\cite{loshchilov2016sgdr}, T-max was 2, and eta was $10^{-5}$.

\subsection{Experiment 1: Can feature extraction networks and association networks learn together well?}
\label{sec:exp1}
The first experiment was conducted to evaluate whether multifeature extraction networks can be combined with AANs from various domains for simultaneous learning.
In this experiment, we used CNNs, which are useful for extracting features.
Further, we selected models that can be expressed through a neurotree based on a convolutional layer and FC layer structures. 
We selected LeNet-5\cite{lecun1989backpropagation} for the image domain, M5\cite{dai2017very} for the sound domain, and CNN\cite{kim-2014-convolutional} for the text domain.

These models were slightly modified and used as $\Psi$($\psi_{image}, \psi_{sound}, \psi_{text}$). Further, we compared and evaluated cases in which they were simultaneously and independently trained.
\begin{figure}[h!]
    \centering
    \includegraphics[height=3.7cm]{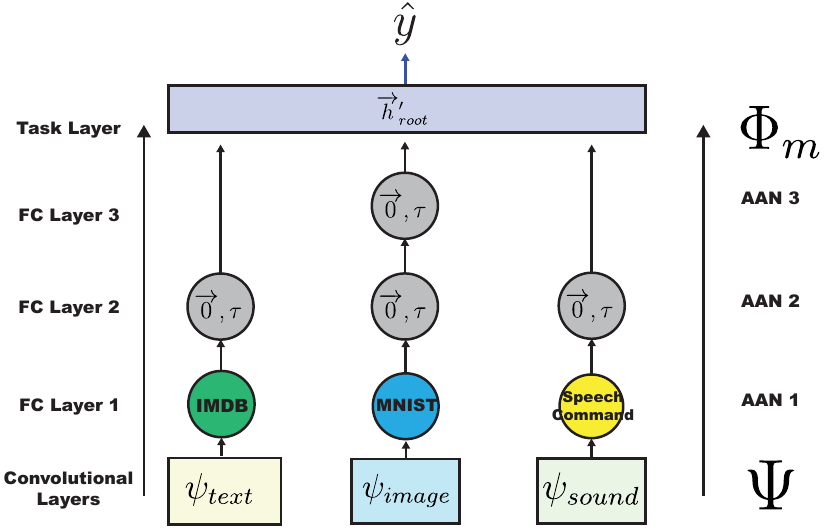}
    \hfil
    \caption{ The neurotrees of Experiment 1.}
\label{fig:img-sound-text-neurotree-ex1}
\end{figure}
LeNet-5\cite{lecun1989backpropagation} was used as the image feature extraction network (in Appendix \ref{table:ex1-img-domain-feature-extractor}).
We created a 128-dimension structure by applying zero padding to the extracted features without using the affine layer (FC layers), and we forwarded it to the AANs.
Batch normalization is not recommended in AANs because the mini-batch size for the domain continues to change during the multifeatured-extraction process (Section \ref{sec:multi-mini-feature-extraction}).
However, in recent image studies, most neural networks use batch normalization; therefore, we used LeNet-5 and it does not use batch normalization.

M5\cite{dai2017very} was used as the sound feature extraction network (in Appendix \ref{table:ex1-sound-domain-feature-extractor}).
We used group normalization\cite{wu2018group} without using batch normalization\cite{ioffe2015batch}; further, we used the affine layer and delivered it to the AANs in 128 dimensions.
This process was implemented in torchaudio\footnote{\url{https://pytorch.org/tutorials/intermediate/speech_command_recognition_with_torchaudio_tutorial.html}}.

A CNN\cite{kim-2014-convolutional} was used as the feature extraction network for the text domain (in Appendix \ref{table:ex1-text-domain-feature-extractor}).

We used the affine layer and delivered it to the AANs in 128 dimensions. We performed classification tasks to compare and evaluate these models more easily. We compared the performance of existing neural networks that can process only one domain with the performance of AANs in which various domains are learned simultaneously when the structures of the neural networks are expressed as a neurotree. 

The AAN parameters used are summarized in Table \ref{table:artificial-association-networks-experiment-models}. The dataset for the image domain was selected as MNIST with 10 classes; the sound dataset included speech commands (SC) with 35 classes, and the text dataset was IMDB with two classes (Table \ref{table:experimental-datasets}).
The neurotree used in this experiment is illustrated in Fig. \ref{fig:img-sound-text-neurotree-ex1}.
\begin{table}[h]
\begin{minipage}{1.0\linewidth}
\centering
\footnotesize
\caption{Test accuracy of learning three domains simultaneously.}
\label{exp:task-47class-results}
\medskip
{\begin{tabular}{l
              c
              @{\hspace*{2.0mm}}c
              @{\hspace*{2.0mm}}c
              c
              @{\hspace*{2.0mm}}c
              @{\hspace*{2.0mm}}c
              }
\toprule
& \multicolumn{3}{c}{For each domain (\%)} 
& \multicolumn{3}{c}{47(=10+35+2) class (\%)}  \\
\cmidrule(lr){2-4}\cmidrule(lr){5-7} 
\footnotesize Model 
& MNIST & SC & IMDB   
& MNIST & SC & IMDB   \\ 
\midrule
LeNet-5    & 98.87 & - & -       
            & 98.87 & - & - 
            \\
M5 (Group Norm)    & - & 92.07 & - 
            & - & 92.07 & - 
            \\
CNN    & - & - & 84.86 
            & - & - & 84.86 
            \\
\midrule
LAN    & 98.63  & 91.26 & 81.87 
       & 98.52  & 91.94  & 82.56   \\    
LAAN   & 98.62 & 91.58 & 81.78  
       & \textbf{98.93} & 91.71 & 82.19   \\
RAN    & 98.69 & 91.65 & 82.39   
       & 98.60  & 91.74  & 82.38   \\     
RAANs  & 98.78 & 90.90 & 82.33  
       & \textbf{98.91} & 91.17   & 82.67   \\
GAU    & 98.73 & 91.59 & 82.32 
       & 98.76  & 91.37  & 82.58   \\     
GAAU  & 98.78 & 91.20 & 82.32  
       & \textbf{98.92} & 91.00   & 82.98   \\       
\midrule
class count  & 10 & 35 & 2
         & 47 & 47 & 47    \\
\bottomrule
\end{tabular}
}    
\end{minipage} 
\end{table}

Convolution is performed if there is no input to a node, and $\overrightarrow{x}'$ is $\overrightarrow{0}$.
\begin{figure*}[ht]
    \centering
    \includegraphics[height=2.9cm]{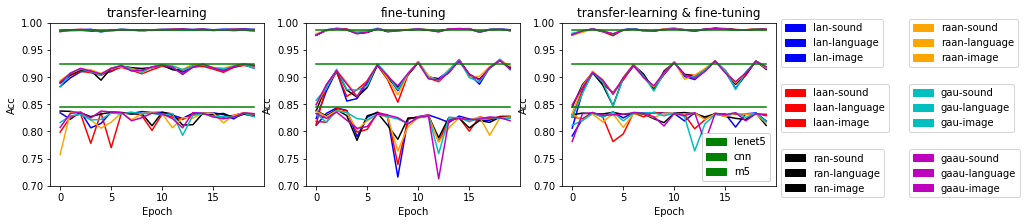}
    \hfil
    \caption{ Test accuracy of transfer learning, fine-turning, and transfer learning with fine tuning.}
\label{fig:experimental-1-fine-neurotree}
\end{figure*}
\begin{table*}[h!]
\hspace{-20px}
\begin{minipage}{1.0\linewidth}
\centering
\caption{Results of Experiment 1 (T denotes transfer learning, F denotes fine-tuning, and the ep denotes the number of training epochs)}
\label{table:task1}
\medskip
{\scriptsize
\begin{tabular}{l
              c
              @{\hspace*{3.0mm}}c
              @{\hspace*{3.0mm}}c
              @{\hspace*{3.0mm}}c
              |c
              @{\hspace*{3.0mm}}c
              @{\hspace*{3.0mm}}c
              @{\hspace*{3.0mm}}c
              c
              @{\hspace*{3.0mm}}c
              @{\hspace*{3.0mm}}c
              @{\hspace*{3.0mm}}c
              |c
              @{\hspace*{3.0mm}}c
              @{\hspace*{3.0mm}}c  
              @{\hspace*{3.0mm}}c
              }
\toprule
& \multicolumn{4}{c}{ \scriptsize{47(=10+35+2) class}}
& \multicolumn{4}{c}{ \scriptsize{Transfer learning} }
& \multicolumn{4}{c}{ \scriptsize{Fine-tuning} }
& \multicolumn{4}{c}{ \scriptsize{Transfer \& Fine-tuning}} 
\\
\cmidrule(lr){2-5}\cmidrule(lr){6-9}\cmidrule(lr){10-13}\cmidrule(lr){14-17}
Model & MNIST & SC  & IMDB & ep & MNIST & SC  & IMDB & ep & MNIST & SC  & IMDB & ep & MNIST & SC  & IMDB & ep\\
\midrule
LeNet-5    & 98.73 & - & - & 30      
            & 98.73 & - & - & 30
            & 98.73 & - & - & 30
            & 98.73 & - & - & 30\\

M5 (Group Norm)    & - & 92.48 & - & 39     
            & - & 92.48 & - & 39
            & - & 92.48 & - & 39
            & - & 92.48 & - & 39\\

CNN    & - & - & 84.50 & 12    
            & - & - & 84.50 & 12
            & - & - & 84.50 & 12
            & - & - & 84.50 & 12\\
\midrule

LAN    & 98.41 & 91.21 & 82.74 & 15    
            & 98.72 & 90.98 & 83.61 & 3
            & \textbf{98.91} & 91.14 & 84.32 & 3
            & \textbf{98.98} & \textbf{92.69} & 83.29 & 19 \\

LAAN  & 98.44 & 90.98 & 82.94 & 11  
            & \textbf{98.77} & 91.09 & 83.22 & 3
            & \textbf{98.97} & 90.90 & 84.27 & 3
            & \textbf{98.90} & \textbf{92.69} & 83.25 & 19 \\

RAN   & 98.14 & 89.00 & 83.35 & 7  
            & 98.61 & 91.29 & 83.45 & 3
            & \textbf{98.88} & 91.27 & 84.00 & 3
            & \textbf{98.87} & \textbf{92.70} & 83.37 & 11 \\

RAAN  & 98.57 & 89.26 & 83.61 & 7  
            & 98.56 & 91.46 & 83.57 & 3
            & \textbf{98.79} & 91.23 & 83.86 & 3
            & \textbf{98.80} & \textbf{92.90} & 83.22 & 19 \\

GAU  & 98.41 & 89.31 & 83.44 & 7  
            & 98.66 & 91.51 & 83.14 & 3
            & \textbf{98.96} & 91.44 & 83.78 & 3
            & \textbf{98.82} & \textbf{92.88} & 83.12 & 15 \\

GAAU  & 98.47 & 89.17 & 83.40 & 7  
            & 98.37 & 90.56 & 83.75 & 5
            & \textbf{98.87} & 91.17 & 83.56 & 3
            & \textbf{98.94} & \textbf{92.95} & 83.30 & 15 \\            

\midrule
pretraining epochs & 0 & 0 & 0 & 
            & 30 & 39 & 12 & 
            & 30 & 39 & 12 & 
            & 30 & 39 & 12 & \\            
pretraining method & - & - & - & 
            & T & T & T & 
            & F & F & F & 
            & F & F & T & \\
\bottomrule
\end{tabular}
}    
\end{minipage} 
\end{table*}
The LAN is equivalent to one FC layer if there are no sibling nodes.
Further, the deepest neurotree level for SC and IMDB was two, while that for MNIST was three.
This implies that the number of convolutions can be different according to the depth of the tree; these different depths can coexist within a single model. This point is not limited to any specific architecture, and this NT can be modified as necessary to accommodate different tasks or complexity levels.

We trained the model to check whether various feature extraction models can be trained simultaneously. Only a word embedding network\cite{pennington2014glove} was used as a pretrained model. 

The test was divided into two cases because the number of classes is different for each dataset.

$\Phi_{m}$ (Classification for each domain): $\Phi_{m}$ was set differently for each domain dataset, and three affine layers were mapped to the class dimension of the datasets $(\phi_{image\_cls}, \phi_{sound\_cls}, \phi_{text\_cls})$.

$\Phi_{m}$ (Classification for 47 classes): One $\Phi_{m}$ was used for all domain datasets; the number of class dimensions was equal to the sum of class dimensions for all datasets. This was done because information on the input domain may be inferred if the last layer is placed differently for each dataset. The purpose here is to break down the boundaries among domains. 

All models were trained over 30 epochs; the results are listed in Table \ref{exp:task-47class-results}(for each domain, 47 classes) and the plots of test accuracy are illustrated in Fig. \ref{fig:experimental-1-neurotree}.
The baseline represents the performance of LeNet-5, M5, and CNN, and this model can be a special case.
The learning processes of AANs are similar to the process of individually learning feature extraction networks, and several problems appear if the model learns in this manner.

LeNet-5 is well trained; however, the M5 requires more training, and the CNN has an overfitting problem. This result indicates that each network has different epochs for optimal performance, and setting learning parameters is more complex than performing individual learning. Thus, an early stopping strategy is difficult to use.
A zigzag pattern is created in the plot of a SC dataset when learning rate scheduling such as cosine annealing is used. Therefore, it is important to determine when to stop for optimal performance.

\subsubsection{Transfer learning and fine tuning}
Therefore, we combined the association model to perform transfer learning and fine tuning after learning feature extraction models individually \cite{pan2009survey}. The result is presented in Table \ref{table:task1} (Transfer learning, Fine tuning).

In transfer learning, the parameters of feature extraction networks are not modified; in fine tuning, the parameters are modified.
We train feature extraction models in the pretraining process and use the validation set to stop learning in the epoch wherein the lowest loss value appears.
Therefore, LeNet-5, M5, and CNN are pretrained with 30, 39, and 12 epochs in 30, 100, and 30 epochs, respectively.

These pretrained models were combined with AANs to perform fine tuning and transfer learning. The networks were re-trained and stopped at the epochs of the lowest validation loss values using the same validation datasets.

The performance was poor in the IMDB dataset in the results of both Table \ref{table:task1}(Transfer learning, Fine-tuning) and Fig. \ref{fig:experimental-1-fine-neurotree}. We believed that this was a result of overfitting because the IMDB dataset was small. Therefore, we reduced the learning rate from 0.001 to 0.0001; however, the results were similar.

Several things can be observed from this plot. First, transfer learning has a slight change in performance, and it does not exceed the performance of existing baseline models compared to fine tuning. However, overfitting is less severe.

By contrast, in the case of fine tuning, the change in performance is large, and it exceeds the performance of the existing baseline model of the LeNet-5, but overfitting is more severe.
Thus, we performed transfer learning for feature extraction models if they were expected to have an overfitting problem; the fine tuning was used for the models with no overfitting problem.

\begin{figure*}[h!]
    \centering
    \includegraphics[height=3.35cm]{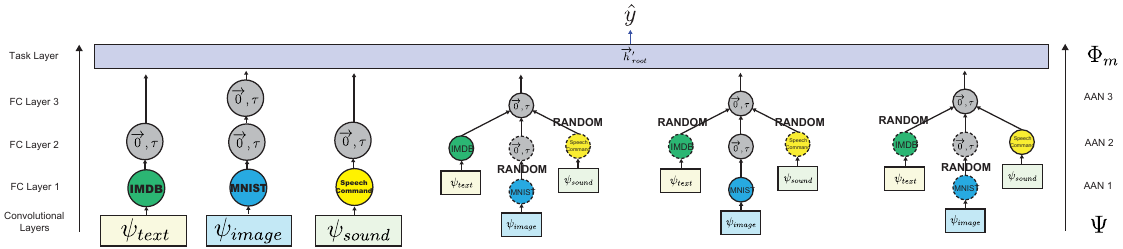}
    \hfil
    \caption{ Neurotree datasets of Experiment 2. }
\label{fig:experimental-2-neurotree}
\end{figure*}

Another problem is that the size of the validation loss among each domain is unequal.
The validation loss of the language model is greater than that of other models.
Thus, we used the validation accuracy as an evaluation score to ensure equal measurements by domain, and we stopped at the epoch with the highest accuracy.

Further, we confirmed that the performance improved on all datasets except for IMDB, where overfitting occurred, as indicated in Table \ref{table:task1}(transfer \& fine-tuning). The network performance can be improved if transfer learning and fine tuning are used by setting the parameters well according to the network.

\subsection{Experiment 2: Can all information be contained?}
\label{sec:exp2}

The information required for the task is obtained along with other additional unnecessary information when we perform a single task in our daily lives. However, we are aware of the other information.

For example, it is possible to recognize the type of music playing in the background while ordering coffee at a cafe; this implies that while performing the task of determining which type of coffee to order, information on music playing can be recognized simultaneously. In other words, humans can always recognize when the information of one or various domains are input.

Thus, the second experiment was conducted to evaluate whether diverse types of information were fully contained in the output vector even if diverse types of information were input simultaneously. 
This experiment focused on whether the output ($\overrightarrow{h}_{root}$) of AANs included embedded information on all data contained in the neurotree. Further, we used the same datasets and models used in Experiment 1.

A dataset was created with neurotrees containing only data from one domain. Data from the image, sound, and text domains were sampled and placed in one neurotree for data augmentation.

The number of combinations for the dataset became too large when the three types of data were combined. 
For example, we need to generate 50,000$\times$80,000$\times$17,500 neurotree samples when 50,000 images, 80,000 sounds, and 17,500 natural language samples are combined.
For this problem, we used a sampling method.
Two additional data items from different domains are randomly sampled when loading each item of data. 
Thus, two neurotree samples are generated from single-domain neurotrees when we load one data item; the resulting neurotree contains three domain data items.
The neurotree samples of the second experiment are illustrated in Fig. \ref{fig:experimental-2-neurotree}.

\begin{table}[h]
\begin{minipage}{0.95\linewidth}
\centering
\caption{Test accuracy of Experiment 2.}
\label{exp:table2}
\medskip
{
\scriptsize
\begin{tabular}{l
              c
              @{\hspace*{2.0mm}}c
              @{\hspace*{2.0mm}}c
              @{\hspace*{2.0mm}}c
              @{\hspace*{2.0mm}}c
              @{\hspace*{2.0mm}}c
              }
\toprule
& \multicolumn{6}{c}{ Task 2: 1 or 3-domain (\%) } \\
\cmidrule(lr){2-7} 
Model & MNIST & MNIST-3D  & SC & SC-3D & IMDB & IMDB-3D \\ 
\midrule
LeNet-5    & 98.87 & - & - & - & - & - \\

M5 (GN)    & - & - & 92.07 & - & - & - \\

CNN    & - & - & - & - & 84.86 & - \\
\midrule
LAN    & 98.66  & 98.48 & 90.53 & 91.45 & 82.70 & 82.64 \\    
LAAN   & 98.72 & 98.68 & 90.73 & 91.31 & 82.40 & 82.28 \\
RAN    & \textbf{98.93} & \textbf{98.91} & 90.41 & 90.57 & 81.46 & 81.08 \\     
RAANs  & \textbf{99.04} & \textbf{98.88} & 90.05 & 91.06 & 83.19 & 83.21 \\
GAU    & 98.85 & 98.75 & 90.99 & 90.67 & 82.32 & 82.19 \\     
GAAU  & 98.85 & 98.76 & 90.67 & 91.55 & 82.17 & 82.01 \\       
       
\bottomrule
\end{tabular}
}    
\end{minipage} 
\end{table}

We constructed the neurotree dataset depicted in Fig. \ref{fig:experimental-2-neurotree} to validate whether the output vector contained all the information in the neurotree; the results are presented in Table \ref{exp:table2}.
The baseline models were the same as those in Experiment 1, with models that could be considered as a special case.

As shown in Table \ref{exp:table2} and the plot (in Appendix. \ref{fig:exp2-all-domain-plot}), the performance decreases slightly in the sound dataset and there is an error margin. However, it becomes increasingly like the plot of the existing model with an increase in the epoch.

In this experiment, there was a smaller error margin when the input data included three domains compared to that when the input data was a single domain for the SC dataset;
this is because the other domains functioned as noise, which makes learning more robust.
These experiments confirmed that various domains of datasets can be learned using one network cell and that multiple types of information can be embedded together. 

\subsection{Experiment 3: Can artificial association networks learn ``deep" neurotrees?}
\label{sec:exp3}
\begin{figure}[h]
    \centering
    \includegraphics[height=3.5cm]{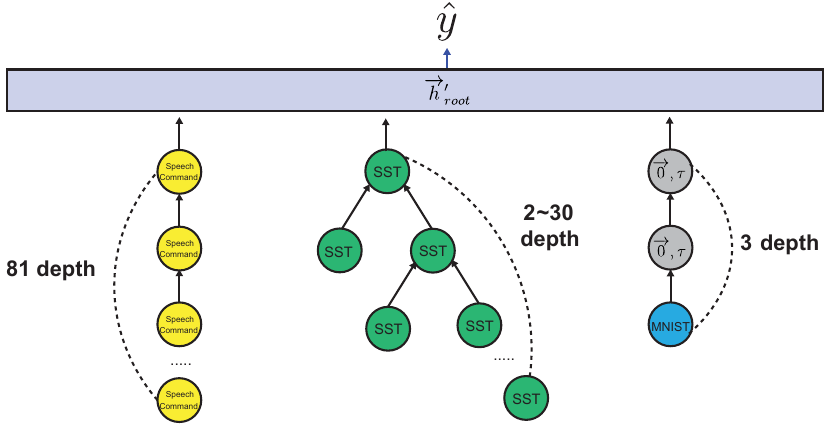}
    \hfil
    \caption{ Tree depths of SC, the Stanford sentiment treebank (SST), and MNIST are 81, 2-30, and 3, respectively. }
\label{fig:img-sound-text-neurotree-ex3}
\end{figure}
This experiment was performed to determine if the networks could be used to train a deep neurotree.
Thus, we utilized the MFCC algorithm for the SC dataset. The baseline models were RNN and GRU models that learn only a single domain dataset (SC dataset).
\begin{figure}[ht]
\begin{floatrow}
\capbtabbox{%
\footnotesize
\begin{tabular}{l
c
              @{\hspace*{5.0mm}}c
              @{\hspace*{5.0mm}}c
              @{\hspace*{5.0mm}}c
              @{\hspace*{5.0mm}}c
              @{\hspace*{5.0mm}}c
              @{\hspace*{5.0mm}}c
              @{\hspace*{5.0mm}}c
}
\toprule
& \multicolumn{3}{c}{ 49(=10+35+2+2) class}
\\
\cmidrule(lr){2-4}
Model & MNIST & SC (MFCC)  & SST \\
\midrule
LeNet-5    & 98.58 & - & - \\
RNN  & - & 62.74 & - \\
GRU    & - & 92.64 & - \\
RAN    & - & - & 72.53 \\
\midrule
LAN    & \textbf{98.84} & 48.60 & 69.10 \\
LAAN  & \textbf{98.80} & 51.49 & 62.58 \\
RAN   & \textbf{98.83} & 3.90 & 52.31 \\
RAAN  & \textbf{98.61} & 5.41 & 52.44 \\
GAU  & \textbf{98.77} & \textbf{93.39} & 70.95  \\
GAAU  & \textbf{98.61} & \textbf{93.23} & \textbf{73.03}  \\
\midrule
class count & 10 & 35 & 2 \\            
\midrule
type & image & sound & text \\
\bottomrule
\end{tabular}
}{%
  \caption{Test accuracy for deep neurotrees.}%
\label{table:img-sound-text-neurotree-ex3}
}
\end{floatrow}
\end{figure}
This experiment is important because the basic RNN model is not well trained with very long-time serial data.
The features were extracted through MFCC by slightly changing Experiment 1: the sound sample rate was set to 16,000; feature size, 40; and  time size, 81.
These data were separated by the time dimension to create a deep neurotree with only one child for each node and a maximum depth of 81 (Fig. \ref{fig:img-sound-text-neurotree-ex3}).

Thus, we generated a neurotree with a depth of 81 through the MFCC algorithm; one FC layer was used in the feature extraction process such that the feature size was from 40 to 128 with ReLU. An FC layer using the same parameters was added to the baseline GRU and RNN models, i.e., an input size of 128 and a hidden size of 128. We proceeded with learning at 30 epochs, and the process is shown in Appendix. \ref{fig:experimental-3-result} and the results in Table \ref{table:img-sound-text-neurotree-ex3}.

The RAN and RAAN models learned well in Experiments 1 and 2;
However, the MNIST dataset (with depth 3) was trained well, but SC (with depth 81) and SST (with depth 2--30) were not learned. When the neurotree was deep, they were even unable to catch up with the performance of the RNN.

The performance of the GAAU model gradually became similar to that of the GRU model even when the tree was deep. Finally, the performance improved.

When learning alone, the RAN trained on the SST dataset but not when learning SC together. In Experiment \ref{sec:exp5}, the RAN was trained for up to 50 epochs.

This result indicates that the complexity of the datasets being learned together affects the learning speed of other datasets.
Further, it is possible that the optimization process has noise if errors are not propagated well when learning with deep sequence datasets.

For LANs and LAANs, the result of overfitting in the language model is shown in the Appendix. \ref{fig:experimental-3-result}.
This is a natural result because the layers of the tree data are learned differently for each level.
Therefore, we proved that learning to express various information delivery structures with deep neurotrees does not significantly affect performance if the error propagates well. The performance of GAAU improved in this experiment.

\subsection{Experiment 4: Can various information be expressed and learned using a neurotree?}
\label{sec:exp4}
\begin{figure}[h]
    \centering
    \includegraphics[height=3.3cm]{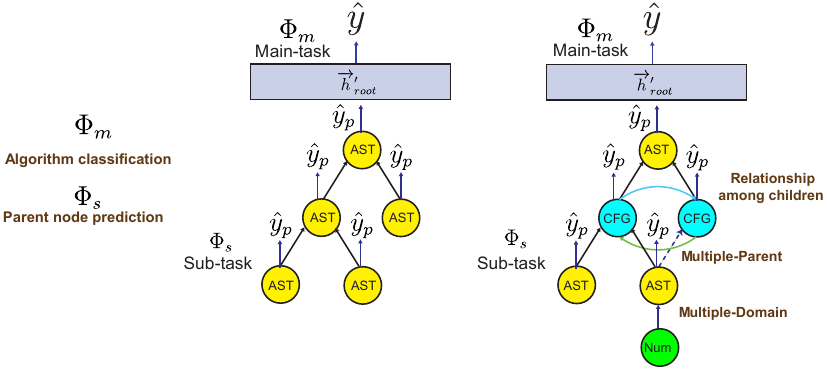}
    \hfil
    \caption{ Abstract syntax tree and neurotree. }
\label{fig:function_neurotree-ex4}
\end{figure}

We explored a domain that can be applied to multidomain, multiparent, relationship, subtask, and main task cases. We selected the source code dataset.
We need to understand the order in which the code is executed to analyze the source code.
Thus, the dataset to be learned must have a main function that generates the first call. We cannot use a dataset in which functions were defined but not called, and therefore, we collected a dataset of source code algorithms for our experiments.

We used a sorting algorithm dataset written in Python from GitHub. It included six algorithm classes of 213 samples (bubble, 43; quick, 46; selection, 38; merge, 47; insert, 25; and shell, 14)\footnote{https://github.com/therk987/Artificial-Association-Networks.git}.
Therefore, using the process introduced in Section \ref{sec:neurotree_architecture}, we represented the source code as a neurotree and compared the performance when learning only the type of information of the AST to that when transforming it into a neurotree (Fig. \ref{fig:function_neurotree-ex4}).

\begin{figure*}[t!]
    \centering
    \includegraphics[height=5.0cm]{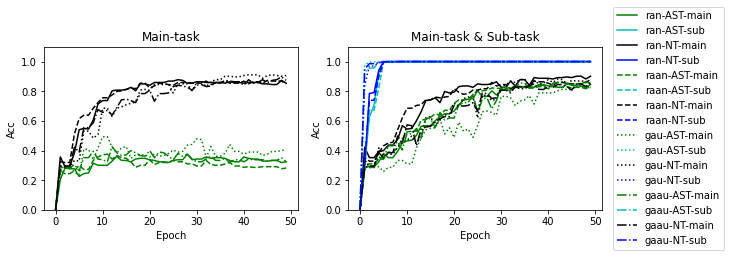}
    \hfil
    \caption{ The left side shows the plot when only the main task is performed, and the right side shows the plot when the main task and subtask are performed together. }
\label{fig:exp4-neurotree-cfg-plot}
\end{figure*}

Further, we embedded the type of each AST node into a vector with 25 dimensions. Consequently, in this experiment, the input size of the AANs was 25 and the size of the hidden state was 128.
Further, the constant value used three FC layers (1, 32), (32, 64), and (64, 25) to represent scale information, and ReLU was applied after each layer. The normalization process was applied only to the last ReLU activation function, and we performed three-fold cross-validation (Table \ref{table:neurotree-ex4}, Appendix. \ref{fig:exp4-neurotree-cfg-plot}) with 50 epochs.
\begin{table}[h]
  \caption{Comparison of AST and neurotree when the subtask is performed or not (test accuracy).}%
  \centering
\footnotesize
  \begin{tabular}{lcc|ccccccc}
\toprule
& \multicolumn{2}{c}{ no subtask }& \multicolumn{4}{c}{ with subtask }\\
\cmidrule(lr){2-7}
& \multicolumn{1}{c}{ AST } & \multicolumn{1}{c}{ Neurotree }& \multicolumn{2}{c}{ AST } & \multicolumn{2}{c}{ Neurotree }\\
\cmidrule(lr){2-7}
Model & $\Phi_{m}$ & $\Phi_{m}$  & $\Phi_{s}$& $\Phi_{m}$ & $\Phi_{s}$ & $\Phi_{m}$ \\
\midrule
RAN    & 32.39 & \textbf{86.85} & 100.0 & 83.57 & 100.0 & \textbf{89.67} \\
RAAN  & 27.70 & \textbf{85.92} & 100.0 & 82.16 & 100.0 & \textbf{84.98} \\
GAU    & 38.50 & \textbf{91.55} & 100.0 & 84.50 & 100.0 & \textbf{87.79} \\
GAAU    & 32.39 & \textbf{86.38} & 100.0 & \textbf{86.85} & 99.99 & 84.97 \\
\midrule
class count & 6 & 6 & 70 & 6 & 70 & 6\\            
\bottomrule
\end{tabular}
\label{table:neurotree-ex4}
\end{table}

We also observed changes in the performance in the first experiment when learning only with the AST node type information and when expressing the dynamic structure using a neurotree, which uses control-flow graphs, constant values, memory store and load operations, function definitions, and call structures.

In the second experiment, the performance was compared by adding a subtask that predicts the parent node in the first experiment.

In addition, if only single-domain information is learned, the relationship information among siblings is not learned, and a single parent node existed. Then, the studied case becomes a RecNN; this special case is used as a baseline.

\begin{table*}[t!]
\begin{minipage}{1.0\linewidth}
\centering
\scriptsize
\caption{Results of Experiment 5 (Could various data structures be simultaneously learned?).}
\label{table:task5}
\medskip
{\begin{tabular}{lcccccccccccccccc}
\toprule

& \multicolumn{3}{c}{\scriptsize Exp1 (\%)} 
& \multicolumn{3}{c}{\scriptsize Exp2 (\%)}  
& \multicolumn{2}{c}{\scriptsize Exp3 (\%)} 
& \multicolumn{1}{c}{\scriptsize Exp4 (\%)} 
& \multicolumn{2}{c}{\scriptsize Exp5 (\%)} \\
\cmidrule(lr){2-4}\cmidrule(lr){5-7}\cmidrule(lr){8-9}\cmidrule(lr){10-10}\cmidrule(lr){11-12}
Model(epoch) & MNIST & SC  & IMDB & MNIST-3D & SC-3D & IMDB-3D & SC-MFCC & SST & Algorithm & UPFD-GOS & Iris \\
\midrule
LeNet-5(15)    & 98.80 & - & - & 98.80 & -& -& -& -& -& -& -\\
M5(35)    & - & \textbf{93.60} & - & - & \textbf{93.60}& -& -& -& -& -& -\\
SCNN(3)    & - & - & \textbf{85.10} & -& -& \textbf{85.10}& -& -& -& -& -\\
RNN(23)    & - & - & - & -& -& -& 55.50& -& -& -& -\\
GRU(23)    & - & - & - & -& -& -& 93.29& -& -& -& -\\
RAN(38)    & - & - & - & -& -& -& -& 72.58 & -& -& -\\
GAU(30)    & - & - & - & -& -& -& -& -& 100.0,90.70 & -& -\\
GAAU(30)    & - & - & - & -& -& -& -& -& 100.0,88.37 & -& -\\
GCN(27)    & - & - & - & - & -& -& -& -& -& 95.50& -\\
GATs(41)    & - & - & - & - & -& -& -& -& -& \textbf{95.74}& -\\
MLP(50)    & - & - & - & - & -& -& -& -& -& -& 100.0\\
\midrule
LAN(43) & 98.73 & 92.27 & 81.85 & 98.58 & 92.86 & 81.92 & 46.05& 69.64& \textbf{100.0},\textbf{93.02}& \textbf{95.82} & 93.33\\
LAAN(47) & \textbf{99.05} & 91.85 & 82.36 & \textbf{98.94} & 93.18& 82.28& \textbf{57.78} & 69.95 & 100.0,83.72& 94.62& \textbf{100.0}\\
RAN(47) & 98.77 & 92.09 & 82.32 & 98.73 & 92.92 & 82.32& \textbf{62.26} & 70.81 & 100.0,86.05 & \textbf{96.21} & \textbf{100.0}\\
RAAN(43) & \textbf{98.84} & 91.91 & 82.70 & 98.68 & 92.72 & 82.62 & \textbf{78.38} & 65.43 & 100.0,83.72 & 95.32 & 93.33\\
GAU(23) & \textbf{98.80} & 91.22 & 82.48 & 98.70 & 92.36 & 82.45 & \textbf{93.42} & \textbf{75.16} & 100.0,88.37 & \textbf{95.74} & \textbf{100.0}\\
GAAU(40) & 98.78 & 91.59 & 81.97 & 98.77 & 91.73 & 81.79& \textbf{93.58} & \textbf{75.29}& \textbf{100.0,88.37} & 95.48 & 93.33\\
\midrule
class count & 10 & 35 & 2 & 10 & 35& 2& 35& 2 & 70,6 & 2 & 3\\            
type & image & sound & text & 3-domain & 3-domain & 3-domain & deep seq & tree & neurotree & graph & tabular\\
\bottomrule
\end{tabular}
}
\end{minipage}
\end{table*}
Consequently, we perform an algorithm classification task as the main task, which uses a subtask to predict the parent node type to determine whether the neurotree, which can express a variety of information, functions correctly.
The model was not fully trained when only the type of the AST node was learned without the subtask. The GAU showed a performance of approximately 91.55\% without performing the subtask by expressing various information through the neurotree.
In addition, the AST node type was learned well when the subtask and main task were performed together, and a performance was achieved when information was expressed as a neurotree.

Although this study is in its initial stages, these results show that even datasets with overly complex structures such as the source code can be learned using diverse types of information. We find and experiment with factors that can improve performance in future research.

\subsection{Experiment 5: Can various data and model structures be learned simultaneously? }
\label{sec:exp5}

Humans can understand various domains and structures using a single brain structure; this experiment was intended to solve the problem of learning about $\exists x$ and $\forall x$ simultaneously using a single model as in an all-in-one approach.
Finally, we simultaneously learn all information using one neural network by using the neurotree dataset we used in the previous experiments.
The performance of this approach was evaluated and compared to that of learning of only a single domain.

In addition, two datasets are added to this experiment: graph (UPFD-GOS\cite{dou2021user}) and tabular (Iris\cite{Dua:2019}) datasets. This is intended to determine whether various data structure datasets can be learned.
For this experiment, if all previous neurotree datasets were learned simultaneously and the performance was not significantly lower than that of the single learning case, this would demonstrate the following: First, the proposed approach can handle various domains and be combined with feature extraction networks such as CNNs. Second, this model may become multimodal and fuse information of various domains. Third, the deep sequence-structure error propagates well. Fourth, it is possible to express complex information structures in a neurotree and perform subtasks simultaneously. Finally, most data structures such as tables, sequences, graphs, and trees can be expressed as a single data structure. This will indicate that it is an all-in-one model.

The dataset included 314,554, 98,119, and 56,667 training, test, and validation samples, respectively (Table \ref{table:experimental-datasets}). It consisted of 11 types of neurotrees containing different domains, data structures, and architectural structure information.

We modified the model slightly to fit the output dimension of the feature extraction process used in the fourth experiment to 128 dimensions. The feature extraction model $\psi_{ast-type}$ added 103-size zero-padding to produce a 128-dimension output; the feature extraction model for the constant value converted the output size of the last layer to 128 dimensions.
The input size of the MLP learning the Iris dataset was 4.
Three layers of (4, 128), (128, 128), and (128, 3), and $L_{2}$ normalization with $10^{-12}$ epsilon and ReLU were used.

$\psi_{tabular}$ in the AAN uses a (4,128) layer and $L_{2}$ normalization with $10^{-12}$ epsilon and ReLU, and a neurotree of depth 3 was used. The GCN and GATs were in the form of an FC layer of (10, 128), and a graph layer of (128, 128) with a FC layer of (128, 2) for the output; LeakyReLU with 0.02 alpha and layer normalization were used.

The $\psi_{graph}$ in AAN used an FC layer of (10, 128) and the depth of the neurotree was 2. We present the test accuracy at the lowest validation loss during 50 epochs in Table \ref{table:task5}; a plot is shown in Appendix. \ref{fig:experimental-5-plot}.

The results of this experiment are similar to those obtained when training occurred separately in Experiments 1, 2, 3, and 4.
Therefore, the results are like those of independent learning even if various datasets are learned simultaneously.
Stop methodologies need to be devised because the number of epochs for optimal performance are different for each model. Thus, it is necessary to use appropriate transfer learning and fine tuning; the parameters must be set well for special cases.

We believe that the results will be similar to that of the special case performance of that model, even if datasets from various domains are learned by one neural network. Further, as shown in Experiment 4, the performance of the model was improved when information from various domains was learned together, or the subtask and main task were performed together.

\section{Discussion and Conclusion}
\subsection{Why is it important to express various routes?}
\begin{table}[h!]
 \caption{Series of association networks.}
  \centering
  \footnotesize
  \begin{tabular}{ll}
    \toprule
    Part     & Concept \\
    \midrule
    
    Entity state & training $\exists x$ \& simultaneous learning for $\forall x$ \\
    Deduction\cite{kim2021deductive} & repeat $\exists x_{t} \oplus \exists x' \rightarrow \exists x_{t+1}$  \\
    Memory\cite{kim2021memory} & memorize for $\exists x$, recall for $\exists x'$\\
    Reinforcement\cite{kim2021imagine} & choose the best memory ($\exists x'$) on the state $x_{t}$\\ & is $x_{t+1}$ what I want?\\
    \bottomrule
  \end{tabular}
  \label{table:concepts}
\end{table}

The disadvantages of the all-in-one models are that setting the learning parameters is more difficult than learning alone; the learning speed is slower; more memory is used than that in GNNs; and overfitting problems occur according to the number of epochs.

The advantage of this type of model is that it is easily scalable to combine various feature extraction models into one neural network; further, it learns with relatively few parameters and in real-world human environments, it can process arbitrary data. Moreover, it is possible to perform batch convolution and model serving such as that provided by BentoML\footnote{This is a model serving framework in MLOps that receives multiple requests through an online service and performs batch convolution, after which it returns the responses to the users. (https://www.bentoml.com)} for various domain data entered when providing online services with neural network models.

The most important advantage of this model can be extended to various studies. The core of this study lies in $f(\exists x) \rightarrow h'_{root}$; thus, a novel humanoid model can be realized if we create a neural network to handle any type of data.
We believe that a human-level recognition system will be created when various data can be recognized, and various tasks can be performed in the environment to realize humanoid-level processing.

Our aim is to create memory\cite{kim2021memory} to memorize $\exists x$ and recall $\exists x'$, which is a deduction\cite{kim2021deductive} that can handle logical expressions repeatedly as $\exists x_{t} \oplus \exists x' \rightarrow \exists x_{t+1}$. Further, we aim to create a system analogous to imagination\cite{kim2021imagine} in which reinforcement learning can move to the desired state through the optimal objective function in the main task process.
Thus, we believe that such a system should be all-in-one, which expresses all structures as one structure even if this incurs a slight performance degradation.

\subsection{Conclusion}
\label{sec:conclusion}
We introduced a data-driven network that can jointly learn relationships and hierarchical information. Simultaneously learning on a variety of datasets can cause performance degradation; however, we showed that this can be done without major degradation. The proposed method does not use a fixed architecture and has been developed to connect diverse types of information being developed. We hope that, in the future, a neural network that can perform all human tasks will be developed.
To this end, we believe that a single embedding device is required to receive data structure inputs from all data generated in the environment.
A further objective of this study is to combine the model with a neural network that can represent the structure of automata and use it with reinforcement learning.
This is an association model that behaves like human sensory organs and can be described to be like a human neural network.
Thus, we believe that this network can be used as a deep neural network component of the DQN\cite{mnih2013playing}.
We will attempt to leverage this network to approach problems that have not been solved before. This paper is part of a series; in the next paper, we will introduce deductive association networks.

  \section*{Acknowledgments}
  Hee-seok Jeong, who has been studying GNN with us in the same laboratory, recommended datasets for the experiments.

\bibliographystyle{IEEEtran}
\bibliography{main.bib}

\appendices
\section{Depth-first search}
\begin{algorithm}[h!]
\caption{Depth-first search.} 
\label{algorithm:dfs}
\begin{algorithmic}[1]
\Function{DFS-left-post-order}{$\mathbf{tree}_{o}$} 
    \State {$N_{o}$ $\gets$ \Call{length}{$\mathbf{C}_{o}$}}
    \For{$i \gets 1...N_{o}$} \label{algoline:leftfirst} \Comment{left-side-first}
        \State {\Call{DFS}{$\mathbf{C}_{o}[i]$}}
    \EndFor
    \State \Call{do-work}{$\mathbf{tree}_{o}$} \label{algoline:postorder} \Comment{post-order}
\EndFunction
\Function{DFS-right-pre-order}{$\mathbf{tree}_{o}$} 
    \State \Call{do-work}{$\mathbf{tree}_{o}$} \label{algoline:preorder} \Comment{pre-order} 
    \State {$N_{o}$ $\gets$ \Call{length}{$\mathbf{C}_{o}$}}
    \For{$i \gets N_{o}...1$} \label{algoline:rightfirst} \Comment{right-side-first}
        \State {\Call{DFS}{$\mathbf{C}_{o}[i]$}}
    \EndFor
\EndFunction
\end{algorithmic}
\end{algorithm}

We decide if we can perform a post-order or pre-order search. Further, it is possible to explore from the left-side-first or from the right-side-first with a bit of correction (1...N or N...1).
Here, $o$ denotes the order of visits, $C_{o}$ denotes the tree's children, and $N_{o} = |\mathbf{C}_{o}|$. The post-order algorithm first navigates through the children using a loop statement and performs do-work to investigate the node, as shown in Fig. \ref{fig:dfs_post}. 
The post-order (and left-side-first) method is used to propagate information from the leaf nodes to the root node.

By contrast, the pre-order algorithm first performs do-work to investigate the corresponding node; then, it navigates through the children using a loop statement, as shown in Fig. \ref{fig:dfs_pre}.

\section{DFC algorithm (Post-order and Left-First)}
\label{appendix:dfc-algorithm}

\begin{algorithm*}[t!]
\caption{Mini-batch DFC} 
\label{algorithm:dfc}
\begin{algorithmic}[1]
\Function{DFC}{$\mathbf{BNN}_{o},lv$} \Comment{mini-batch neuronodes}
    \State{$\mathbf{BNN}_{o}.visit\_counts \gets \mathbf{BNN}_{o}.visit\_counts + 1$} \label{algoline:visit_count} \Comment{for deconvolutional propagation}
    \State{$\mathbf{BNN'}_{o} \gets \mathbf{BNN}_{o}.get\_not\_calculated\_nodes()$} \label{algoline:notvisited} \Comment{haven't visited yet} 
    \If{ $\mathbf{BNN'}_{o}.is\_empty()$ } 
        \State{\Return} \label{algoline:no_neuronodes} \Comment{no nodes need to calculate}
    \EndIf 
    \State {$\mathbf{x}_{o},\mathbf{\tau}_{do},\mathbf{\tau}_{so},\mathbf{BA}_{co},\mathbf{BC}_{o}$ $\gets$ $\mathbf{BNN'}_{o}.items()$} \label{algoline:getitems}
    \State{$N_{o} \gets max(\mathbf{BC}_{o}.get\_counts())$} \label{algoline:maximum_child_count} \Comment{maximum child count}
    \For{$i \gets 1...N_{o}$} \Comment{left-side-first}
        \State{$\mathbf{BC}'_{o}[i] \gets \mathbf{BC}_{o}.get(order=i)$}
        \State {\Call{DFC}{$\mathbf{BC}'_{o}[i],lv+1$}}  \label{algoline:post-order-conv} \Comment{post-order}
    \EndFor 
    \State {$\overrightarrow{\mathbf{x}}'_{o}, *info_{1}$ $\gets$ $\Psi[\tau_{do}]({\mathbf{x}}_{o})$} \label{algoline:multi_feature_extraction} \Comment{feature extraction}
    \If{$N_{o}$ is $0$} \Comment{leaf node} \label{algoline:graph_aggregate}
        \State {$\overrightarrow{\mathbf{h}}_{o} \gets \overrightarrow{0}$}
        
    \Else
        \State {$\mathbf{h}_{o}$ $\gets$ $\mathbf{BC}_{o}.get\_hiddens()$} \label{algoline:get_child_hidden} \Comment{child node hidden states have already been stored}
        \State {$\overrightarrow{\mathbf{h}}_{o}, info_{2}$ $\gets$ $g(\mathbf{GNN}(\mathbf{BA}_{co}, \mathbf{h}_{o}))$} \Comment{$g(\mathbf{A}_{co}\mathbf{h}_{o})$}
    \EndIf \label{algoline:graph_aggregate2}
    
    \State {$\overrightarrow{\mathbf{h}'}_{o}$ $\gets$ $\mathbf{RNN}$($\overrightarrow{\mathbf{x}}'_{o},\overrightarrow{\mathbf{h}}_{o}$)} \label{algoline:recurrentconv} \Comment{$\mathbf{W}[\overrightarrow{x}_{o}, \overrightarrow{h}_{o}]$}   
    
    \State {$\hat{\mathbf{y}}_{o}$ $\gets$ $\Phi_{s}[\tau_{to}]$($\overrightarrow{\mathbf{h'}}_{o}$)} \label{algoline:do_tasks} \Comment{do subtasks} 
    \State {$\mathbf{BNN'}_{o}.\overrightarrow{\mathbf{h}'} \gets \overrightarrow{\mathbf{h}'}_{o}$} \label{algoline:store_hiddens} \Comment{store hidden states}
    \State {$\mathbf{BNN'}_{o}.\hat{\mathbf{y}} \gets \hat{\mathbf{y}}_{o}$} \label{algoline:store_outputs} \Comment{store the outputs of tasks}
    \State {$\mathbf{BNN'}_{o}.more$ $\gets$ $(*info_{1}, *info_{2})$} \label{algoline:store_infomation} \Comment{store more information}
    \State \Return
    \EndFunction
\end{algorithmic}
\end{algorithm*}

\begin{figure}[ht]
    \centering
    \includegraphics[height=5.0cm]{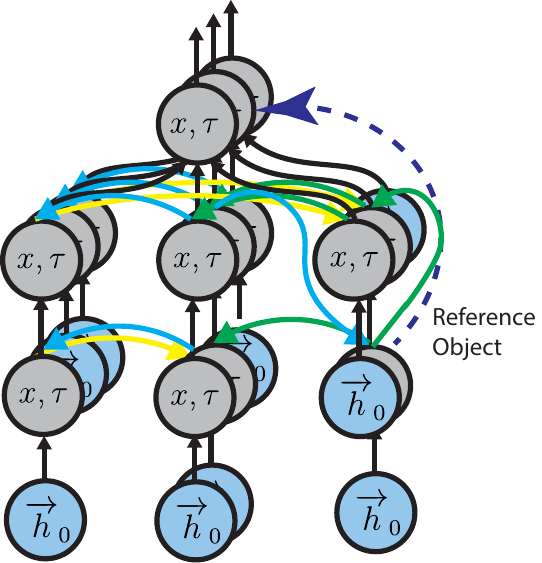}
    \hfil
    \caption{ Batch DFC. }
\label{fig:batch-dfc}
\end{figure}

The input of this algorithm is a neurotree of the mini-batch structure ($\mathbf{BNN}_{o}$) described as an algorithm for the learning process (Algorithm \ref{algorithm:dfc}).
We describe the recursive algorithms for convolution performed in \textbf{the neuronode of the o-th order in the neurotree} in mini-batch form.
An example of that convolution order o is shown in Fig. \ref{fig:dfc}, \ref{fig:batch-dfc}.

First, the number of visits must be calculated when visiting each node to tour neuronodes that are not visited only(lines \ref{algoline:visit_count}--\ref{algoline:no_neuronodes}).
A neuronode can be a child node of multiple nodes because the neurotree has multiple parent nodes. Thus, a neuronode that has already been toured can be toured again as a child node.

To prevent this, only neuronodes visited for the first time in the mini-batch ($\mathbf{BNN}_{o}$) are used as input for the current node ($\mathbf{BNN'}_{o}$). 
The previously calculated result is reused because the hidden state has been calculated for the visited neuronodes; the recalculation is not performed.
Further, it is possible to perform each calculation only once in the DFD process (Algorithm \ref{algorithm:dfd}) by using the number of visits.

Second, we obtain the five elements ($x_{o}, \tau_{do}, \tau_{to}, \mathbf{A}_{co}, \mathbf{C}_{o}$) of the neuronodes defined in Section \ref{sec:neuro_datastructure} in the mini-batch form (line \ref{algoline:getitems}).
Here, $o$, $\mathbf{x}_{o}$, $\tau_{do}$, $\tau_{to}$, and $\mathbf{BA}_{co}$ and $\mathbf{BC}_{o}$ represent the convolution order of the DFC, input data of the corresponding nodes, domain information of the corresponding nodes, subtask to be performed on the corresponding nodes, and relationship information ($\mathbf{A}_{co}$) between children and their children ($\mathbf{C}_{o}$) in the mini-batch form, respectively.

Third, we recursively tour child nodes ($\mathbf{BC}_{o}$) using a left-side-first search in the post-order form (lines \ref{algoline:maximum_child_count} $\sim$ \ref{algoline:post-order-conv}). As $\mathbf{BC}_{o}$ has the sample size of the mini batch, the number of children ($\mathbf{C}_{o}$) vary for each sample. The largest number of children is called $N_{o}$.

Children present at $i$ are called $\mathbf{BC}'_{o}[i]$ if the order from 1 to $N_{o}$ is $i$; then, the DFC is performed again on these nodes to traverse them recursively.
The DFC will perform the propagation in the order of DFS (Section \ref{sec:related_works_dfs}) using post-order and left-side-first structures because of this process.

Thus, $o$ becomes the convolution order of the propagation path from the leaf node to the root node (Fig \ref{fig:dfc}.

Fourth, we also introduce the process of extracting the features of the data $\mathbf{x}_{o}$ of the corresponding neuronodes (line \ref{algoline:multi_feature_extraction}).
Further, $\Psi$ denotes multifeature-extraction networks (Section \ref{sec:multi_feature_extraction}) implemented in a dictionary data structure. The domain ($\tau_{do}$) and input data ($\mathbf{x}_{o}$) at the current node are inputs of $\Psi$. The feature extraction networks corresponding to the domain are selected by entering $\tau_{do}$ as a key value; the input data $\mathbf{x}_{o}$ is converted into feature vectors ($\overrightarrow{\mathbf{x}}_{o}’$). This process is grouped by the domain, and the features are extracted into the multi-mini-batch structure. This method is described in Algorithm \ref{algorithm:batch_type_embedding}.

Fifth, we introduce the process of aggregating information in the child nodes of the o-th neuronodes with the relation terms (lines \ref{algoline:graph_aggregate} $\sim$ \ref{algoline:graph_aggregate2}).

The children are already visited in the post order when this process is executed.
Thus, the hidden states of children have already been calculated and stored in their neuronodes; the hidden states can be obtained.
For leaf nodes, $\overrightarrow {0}$ is used because there are no more children. Meanwhile, each piece of information received by the children is an entity. Further, the entities of the child node have relationships with each other, and we can express this relationship as a graph ($\mathbf{A}_{c}$). 
The convolution process is structured similar to a graph processing layer such as that in GCNs, GATs, and EGNNs\cite{kipf2016semi, velivckovic2017graph, gong2019exploiting}. 
We aggregate the convolution outputs as a readout process to obtain $\overrightarrow{\mathbf{h}}_{o}$. This process makes it possible for us to express hierarchical information as a tree and relational information as information in a graph.
We use this process as if we were learning the relationships between sibling nodes.

Sixth, the output of the hidden state of the current neuronode is obtained (line \ref{algoline:recurrentconv}).
A convolution is performed using the form of an RNN with $\overrightarrow{\mathbf{x}}'_{o}$ obtained in the fourth process and $\overrightarrow{\mathbf{h}}_{o}$ aggregating the information of the children in the fifth process. Finally, the output ($\overrightarrow{\mathbf{h}'}$) represents the current hidden state ($\overrightarrow{\mathbf{h}}' \gets \mathbf{RNN}(\overrightarrow{\mathbf{x}}', \overrightarrow{\mathbf{h}})$).
This process can function as an GRU\cite{chung2014empirical}.

Seventh, the subtask is performed with the hidden state ($\overrightarrow{\mathbf{h}}'$) obtained in the sixth step (line \ref{algoline:do_tasks}.
This process makes it possible to perform tasks at the node level.
The appropriate subtask is the task of predicting the parent node (Section \ref{sec:multi_task}).

Finally, the hidden state (6th) and output of the subtask (7th) are stored in the current neuronode (lines \ref{algoline:store_hiddens} $\sim$ \ref{algoline:store_infomation}).
This process makes it possible to reuse the already calculated hidden state (5th process) and deliver the hidden state to the multiparent node.
Further, maxpool is used in the aggregation ($g$) process; it is possible to store the index information and use it in the deconvolution process to perform unpooling (line \ref{algoline:store_infomation}).

The $\overrightarrow{h'}_{root}$ of the root neuronode is finally obtained by repeating this process, and batch learning is performed by performing only one set of calculations in each node from the leaf node to the root node. Thus, a neurotree is obtained, and we contain the hidden state and subtask output in each node. Implementing this recursive function as a loop (while) will increase the speed.

\section{DFD (Pre-order and Right-First)}
\label{appendix:dfd}
The DFD was introduced as a propagation methodology to be extended to an autoencoder or bidirectional model. This study shows that the structures of existing neural network models can be expressed in data (as a neurotree) and learned.
This is a propagation methodology for traversing and restoring DFCs in the reverse order, and for expressing multiple parent structures and restoring them.
Thus, if the DFC travels from leaf nodes to root nodes and propagates, the DFD is an algorithm that travels from the root nodes to leaf nodes and propagates. An example of this deconvolution order o is shown in Fig. \ref{fig:dfd}, \ref{fig:batch-dfd}. Further, we describe the deconvolution process in Appendix \ref{appendix:dfd}.
\begin{algorithm*}[t!]
\caption{Mini-batch DFD.}
\label{algorithm:dfd}
\begin{algorithmic}[1]
    \Function{DFD}{$\protect\overrightarrow{\mathbf{dx}}_{o},\protect\overrightarrow{\mathbf{dh}}_{o},\mathbf{BNN}_{o},*lv$}
        \State {$\mathbf{BNN}_{o}.\overrightarrow{\mathbf{dx}} \gets \mathbf{BNN}_{o}.\overrightarrow{\mathbf{dx}} + \overrightarrow{\mathbf{dx}}_{o}$} \label{algoline:add_x}
        \State {$\mathbf{BNN}_{o}.\overrightarrow{\mathbf{dh}} \gets \mathbf{BNN}_{o}.\overrightarrow{\mathbf{dh}} + \overrightarrow{\mathbf{dh}}_{o}$} \label{algoline:add_h}

        \State{$\mathbf{BNN}_{o}.dvisit\_counts \gets \mathbf{BNN}_{o}.dvisit\_counts + 1$} \label{algoline:dvisit_count} \Comment{counting}
        \State{$\mathbf{BNN'}_{o} \gets \mathbf{BNN}_{o}.get\_final\_visit\_nodes()$} \label{algoline:final_visit} \Comment{if dconv.count == conv.count} 
        \If{ $\mathbf{BNN'}_{o}.is\_empty()$ } 
        \State \Return \label{algoline:dconv_no_neuronodes}
        \EndIf 
    
        \State {$\overrightarrow{\mathbf{dx}_{o}}$ $\gets$ $\mathbf{BNN'}_{o}.\overrightarrow{\mathbf{dx}_{o}}/\mathbf{BNN'}_{o}.dvisit\_counts$} \label{algoline:div_x}
        \State {$\overrightarrow{\mathbf{dh}_{o}}$ $\gets$ $\mathbf{BNN'}_{o}..\overrightarrow{\mathbf{dh}_{o}}/\mathbf{BNN'}_{o}.dvisit\_counts$} \label{algoline:div_h}
        \State {$\overrightarrow{\mathbf{dx}}'_{o},\overrightarrow{\mathbf{dh}}'_{o}$ $\gets$ $\mathbf{RNN}^{-1}$($\overrightarrow{\mathbf{dx}_{o}}, \overrightarrow{\mathbf{dh}}_{o}$)} \label{algoline:deconv_recurrent} \Comment{$\mathbf{W}(\overrightarrow{x}_{o},\overrightarrow{h}_{o})$}    
        
        \State {$\mathbf{\tau}_{do},\mathbf{BA}_{co},\mathbf{BC}_{o}$ $\gets$ $\mathbf{BNN'}_{o}.items()$} \label{algoline:deconv_getitems}
        \State $*info_{1}, *info_{2}$ $\gets$ $\mathbf{BNN}'_{o}.more$
        \State $\mathbf{BNN}_{o}.\hat{\mathbf{x}}_{o}$ $\gets$ $\Psi^{-1}[\tau_{do}](\overrightarrow{\mathbf{dx'}}_{o}, *info_{1})$ \label{algoline:multi_feature_decoding}
        \State{$N_{o} \gets max(\mathbf{BC}_{o}.get\_counts())$} \label{algoline:d_maximum_child_count} \Comment{maximum child count}
        \If{$N_{o}$ is $0$} \Comment{leaf node}
        \State \Return \label{algoline:deconv_return}
        \EndIf
        \State $\tilde{\mathbf{dh}}_{o} \gets g^{-1}(\overrightarrow{\mathbf{dh}}'_{o},*i_{2})$
        \State $\tilde{\mathbf{dh}}'_{o}$ $\gets$ $\mathbf{GNN}^{-1}(\mathbf{BA}_{co},\tilde{\mathbf{dh}}_{o},lv)$ \label{algoline:deconv_restore}
        \For{$i \gets N_{o}...1$} \label{algoline:deconv_rightfirst} \Comment{right-side-first}
            \State {$\mathbf{BC}'_{o}[i], *batch\_indices \gets \mathbf{BC}_{o}.get(order=i)$}
            \State {\Call{DFD}{$\protect\overrightarrow{\protect\mathbf{dx}}'_{o}[batch\_indices], \protect\tilde{\mathbf{dh}}'_{o}[batch\_indices,i], \protect\mathbf{BC}'_{o}[i], lv+1$}}
        \EndFor \label{algoline:deconv_rightfirst_loop}

        \State \Return
    \EndFunction
\end{algorithmic}
\end{algorithm*}
The first step is to add $\overrightarrow{\mathbf{dx}}_{o}$ and $\overrightarrow{\mathbf{dh}}_{o}$ received from the parent node to the $\overrightarrow{\mathbf{dx}}_{o}$ and $\overrightarrow{\mathbf{dh}}_{o}$ of the current neuronode (line \ref{algoline:add_x}$\sim$\ref{algoline:add_h}).
Similar to DFC, we count how many times we have visited the current node (line \ref{algoline:dvisit_count}).
This is done because the neuronode can have multiple parent nodes, so the hidden states of parent nodes are delivered to the current neuronode several times.

Second, we only bring neuronodes ($\mathbf{BNN'}_{o}$) when the number of visits in the DFC process and that in the DFD process are equal (line \ref{algoline:final_visit}). 
This denotes the last time of the visit in which the current node received information from all parent nodes.

Third, at the time of the last visit, the inputs and hidden states received from parents are averaged by the number of visits, and the result is used as input for RNNs.
In addition, this study was inspired by the existing LSTM autoencoder; we used the information in which parents predicted their feature vector. 

The average $\overrightarrow{\mathbf{dx}}_{o}$ and $\overrightarrow{\mathbf{dh}}_{o}$ are delivered from the parent nodes to the current node and used as input for RNNs.
This process allows the system to receive and learn information with multiple parent nodes in the deconvolution process.

Fourth, the items needed during the deconvolution process are removed, and the items are utilized to restore them through multifeature restoration networks ($\Psi^{-1}$) corresponding to multifeature extraction networks ($\Psi$). This process is similar to the relationship between the encoder and decoder. Through this process, the predicted $\hat{\mathbf{x}}$ is obtained and stored in the neuronode.

Fifth, if a child exists in the neuronode, the hidden state to be delivered to the children is restored; otherwise, it is returned.

\begin{figure}[ht]
    \centering
    \includegraphics[height=5.5cm]{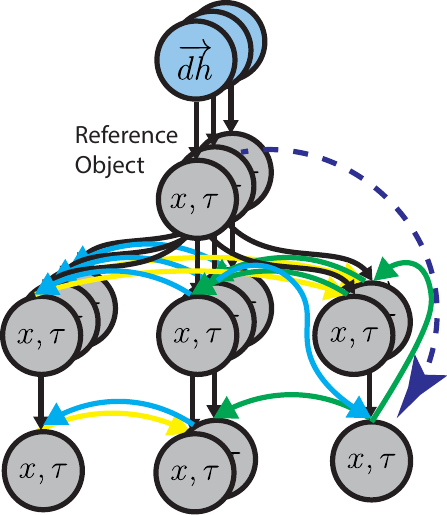}
    \hfil
    \caption{ Batch DFD. }
\label{fig:batch-dfd}
\end{figure}

The author proceeded with the aggregate process in DFC through maxpool, saved the indices, and restored them in DFD using the unmaxpool method (line \ref{algoline:d_maximum_child_count}$\sim$\ref{algoline:deconv_restore}).

Finally, we recursively tour the child nodes in reverse order ($N_{o}...1$).
$N_{o}$ is the maximum number of children among the mini-batch samples.

At this time, the mini-batch in which the i-th child exists is recursively delivered to the children $\overrightarrow{\mathbf{dx}}'_{o}[batch\_indices]$ and $\tilde{\mathbf{dh}}'_{o}[batch\_indices]$, and deconvolution is performed through DFD.

This process propagates in pre-order and right-side-first order, and batch learning can be achieved by computing nodes of all trees only once (line \ref{algoline:deconv_rightfirst}$\sim$\ref{algoline:deconv_rightfirst_loop}).

\section{Multi-Batch Feature Extraction Algorithm}
\label{appendix:multi-mini-feature-extraction}
\begin{algorithm*}[t!]
\caption{Multi-batch feature extraction method.}
\label{algorithm:batch_type_embedding}
\begin{algorithmic}[1]
\Function{multi-batch-convolution}{$\mathbf{BNN}_{o}$}
        \State $dict_{x}$ = \{\} \label{algoline:x_dict}\Comment{key: $\tau$, default\ value: $list_{x}$}
        \State $dict_{idx}$ = \{\} \label{algoline:idx_dict}\Comment{key: $batch\_idx$, value: $(\tau_{d}$, $index_{dict_{x}})$}
        \State $B_{o}$ = $\mathbf{BNN}_{o}$.size() \label{algoline:batch_size} \Comment{mini-batch-size}
        \For{$idx_{batch} \gets 1...B_{o}$} \label{algoline:batch_loop} \Comment{Grouping}
            \State $\mathbf{NN}_{o}$ = $\mathbf{BNN}_{o}[idx_{batch}]$ \label{algoline:idx_batch_item}
            \State $dict_{idx}$[$idx_{batch}$] = ($\mathbf{NN}_{o}.\tau_{d}$, len($dict_{x}$[$\mathbf{NN}_{o}.\tau_{d}$])) \label{algoline:save_idx} \Comment{store the location of the mini-batch} 
            \State $dict_{x}$[$\mathbf{NN}_{o}.\tau_{d}$].append($\mathbf{NN}_{o}.x$) \label{algoline:grouping_x} \Comment{grouping $x$ with the same domain}
        \EndFor
        \For{$\tau_{d}, \mathbf{x}_{\tau_{d}}$ $\gets$ $dict_{x}.items()$} \label{algoline:grouping_conv_loop} \Comment{$\tau_{d}(key),\mathbf{x}(values)$}
        \State {$\overrightarrow{\mathbf{x}}_{\tau_{d}}$ = $\Psi$[$\tau_{d}$]($\mathbf{x}_{\tau_{d}}$)} \label{algoline:grouping_conv} \Comment{Multi-mini-batch convolution} 
        \State {$\overrightarrow{\mathbf{x}}'_{\tau_{d}}$ = [$\overrightarrow{\mathbf{x}}_{\tau_{d}}$,onehot($\tau_{d}$).repeat(len($\mathbf{x}_{\tau_{d}}$))]} \label{algoline:grouping_concat_bias} \Comment{Domain bias}
        \State {$dict_{x}$[$\tau_{d}$] = $\overrightarrow{\mathbf{x}}'_{\tau_{d}}$} \label{algoline:grouping_store}
        \EndFor
        \State $output_{batch}$ = [] \label{algoline:restore_batch_list} \Comment{Ungrouping}
        
        \For{$idx_{batch}$ $\gets$ $1...B_{o}$} \label{algoline:restore_batch_loop} \Comment{restore the location of the mini-batch}
            \State {$\tau_{d}, idx_{\tau_{d}}$ $\gets$ $dict_{idx}$[$idx_{batch}$]} \label{algoline:restore_batch_idx_get}
            \State {$\overrightarrow{\mathbf{x}}'_{\tau_{d}}$ = $dict_{x}$[$\tau_{d}$]} \label{algoline:restore_batch_get}
            \State $output_{batch}$.append($\overrightarrow{\mathbf{x}}'_{\tau_{d}}[idx_{\tau_{d}}]$) \label{algoline:restore_batch_output}
        \EndFor
        \State \Return $Stack(output_{batch})$ \label{algoline:restore_batch_return}
\EndFunction
\end{algorithmic}
\end{algorithm*}

First, this algorithm creates two dictionary data structures; the default data type of $dict_{x}$ is a list (e.g., defaultdict(list) in Python). $dict_{x}$ represents a dictionary that groups x for each domain, and $dict_{idx}$ represents a dictionary for storing batch indices at which grouped input x exists (line \ref{algoline:x_dict}$\sim$\ref{algoline:idx_dict}). 

We tour the mini-batch samples in order, and store the batch index ($idx_{batch}$) and group data ($x_{\tau_{d}}$) with the same domain ($\tau_{d}$)(line \ref{algoline:batch_loop}$\sim$\ref{algoline:grouping_x}). Therefore, the inputs are collected for each domain, which is called $\mathbf{x}_{\tau_{d}}$.

Mini-batch samples ($\mathbf{x}_{\tau_{d}}$) of domains are used as inputs of multifeature extraction networks ($\Psi$) that perform convolution with feature extraction networks ($\psi_{\tau_{d}}$) that correspond to domains.
We concatenate the domain-bias vector (Section \ref{sec:multi_feature_extraction}) to obtain $\overrightarrow{\mathbf{x}}_{\tau_{d}}'$(line \ref{algoline:grouping_conv_loop}$\sim$\ref{algoline:grouping_store}).

Finally, we restore the original batch index using $dict_{idx}$ to ungroup (line \ref{algoline:restore_batch_list}$\sim$\ref{algoline:restore_batch_return}) to release a group.

\section{Feature extraction models}

\begin{table}[ht]
\begin{floatrow}
\capbtabbox{%
\footnotesize
\begin{tabular}{l
              c
              @{\hspace*{2.0mm}}c
              @{\hspace*{2.0mm}}c
              @{\hspace*{2.0mm}}c
              @{\hspace*{2.0mm}}c
              }
\toprule
& \multicolumn{5}{c}{ $\psi_{image}$ based on LeNet-5\cite{lecun1989backpropagation}}  \\
\cmidrule(lr){2-6}
\scriptsize Layer 
& In & Out  & Kernel & Stride & $\sigma$    \\ 
\midrule
Conv2D    & 1  & 6 & (5,5) & 1 & tanh  \\    
AvgPool2D & - & - & (2,2) & 1 & - \\
Conv2D    & 6 & 16 & (5,5) & 1 & tanh  \\     
AvgPool2D & - & - & (2,2) & 1 & -   \\
Conv2D    & 16 & 120 & (5,5) & 1 & tanh  \\     
padding & 120  & 128 & - & - & -   \\
\midrule
Final    & - & 128 & - & - & - \\
\bottomrule
\end{tabular}
}{%
  \caption{$\psi_{image}$ for image domain.}%
\label{table:ex1-img-domain-feature-extractor}
  
}

\end{floatrow}
\end{table}

\begin{table}[ht]
\begin{floatrow}
\capbtabbox{%
\footnotesize
\begin{tabular}{l
              c
              @{\hspace*{2.0mm}}c
              @{\hspace*{2.0mm}}c
              @{\hspace*{2.0mm}}c
              @{\hspace*{2.0mm}}c
              }
\toprule
& \multicolumn{5}{c}{ $\psi_{text}$ based on CNN\cite{kim-2014-convolutional} }  \\
\cmidrule(lr){2-6}
\scriptsize Layer 
& In & Out  & Kernel & Stride & $\sigma$    \\ 
\midrule
Conv2D    & 1  & 100 & (3,3) & 1 & relu  \\    
AvgPool2D & - & - & (2,2) & 1 & - \\
Conv2D    & 1 & 100 & (4,4) & 1 & relu  \\     
AvgPool2D & - & - & (2,2) & 1 & -   \\
Conv2D    & 1 & 100 & (5,5) & 1 & relu  \\     

Concat  & (300,400,500)  & 1200 & - & - & -   \\

Dropout  & -  & - & - & - & -   \\
FC layer  & 1200 & 128   & - & - & -   \\
\midrule
Final  & - & 128 & - & - & - \\
\bottomrule
\end{tabular}
}{%
  \caption{$\psi_{text}$ for text domain.}%
  \label{table:ex1-text-domain-feature-extractor}
}
\end{floatrow}
\end{table}

\begin{figure}[ht]
\begin{floatrow}

\capbtabbox{%
\footnotesize
\begin{tabular}{l
              c
              @{\hspace*{2.0mm}}c
              @{\hspace*{2.0mm}}c
              @{\hspace*{2.0mm}}c
              @{\hspace*{2.0mm}}c
              @{\hspace*{2.0mm}}c
              }
\toprule
& \multicolumn{6}{c}{ $\psi_{sound}$ based on M5\cite{dai2017very} }  \\
\cmidrule(lr){2-7}
\scriptsize Layer 
& In & Out  & Kernel & Stride & Norm & $\sigma$    \\ 
\midrule 
Conv1D    & 1  & 128 & 80 & 4 & group 16 & relu \\    
MaxPool1D & - & - & 4 & 1 & -  & -  \\
Conv1D    & 128 & 128 & 3 & 1 & group 16 & relu   \\     
MaxPool1D & - & - & 4 & 1 & -  & -    \\
Conv1D    & 128 & 256 & 3 & 1 & group 16 & relu  \\     
MaxPool1D & - & - & 4 & 1 & -  & -    \\
Conv1D    & 256 & 512 & 3 & 1 & group 16 & relu  \\     
MaxPool1D & - & - & 4 & 1 & -  & -    \\
AdaptiveAvgPool & - & 1 & - & - & -  & -  \\
FC layer & 512 & 128   & - & - & -  & relu   \\
\midrule
Final & - & 128 & - & - & -  & - \\
\bottomrule
\end{tabular}
}{%
  \caption{$\psi_{sound}$ for sound domain.}%
\label{table:ex1-sound-domain-feature-extractor}
}
\end{floatrow}
\end{figure}

\section{Test accuracy plot of the experimental 2}
\begin{figure*}[t]
    \centering
    \includegraphics[height=9cm]{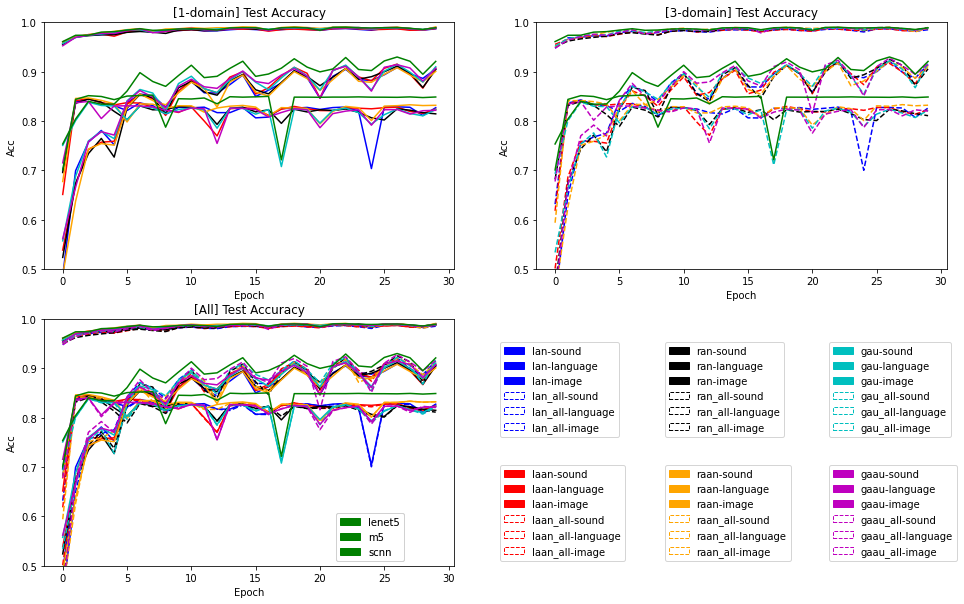}
    \hfil
    \caption{We divided the test accuracy plots shown in Experiment 2 into three plots.}
\label{fig:exp2-all-domain-plot}
\end{figure*}

\section{Test accuracy plot of the experimental 3}
\begin{figure*}[t!]
    \centering
    \includegraphics[height=9cm]{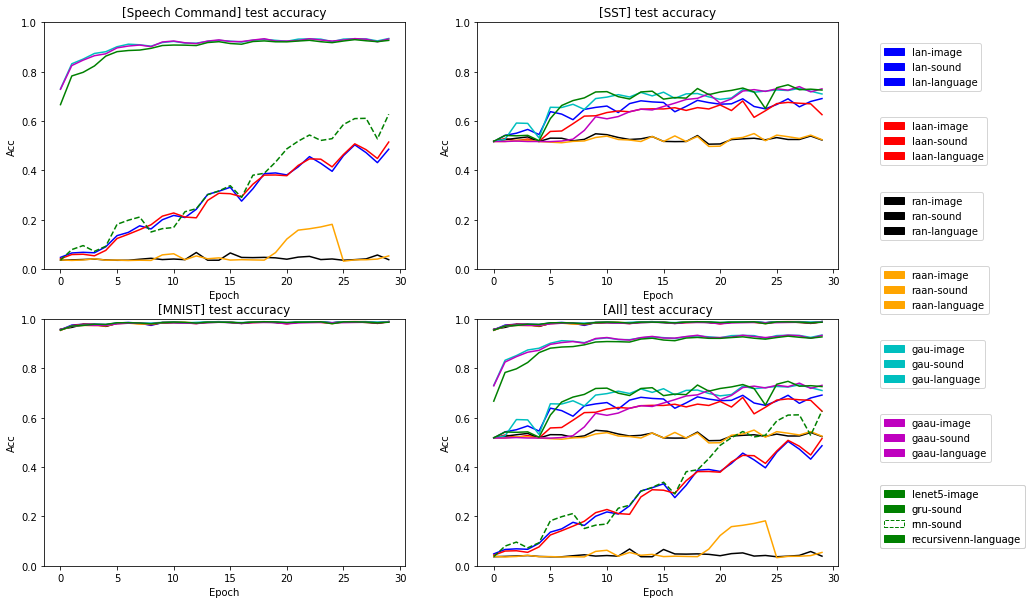}
    \hfil
    \caption{We divided the test accuracy plots shown in Experiment 3 into three plots.}
\label{fig:experimental-3-result}
\end{figure*}

\section{Test accuracy plot of the experimental 4}

\section{Test accuracy plot of the experimental 5}
\begin{figure*}[t!]
    \centering
    \includegraphics[height=20cm]{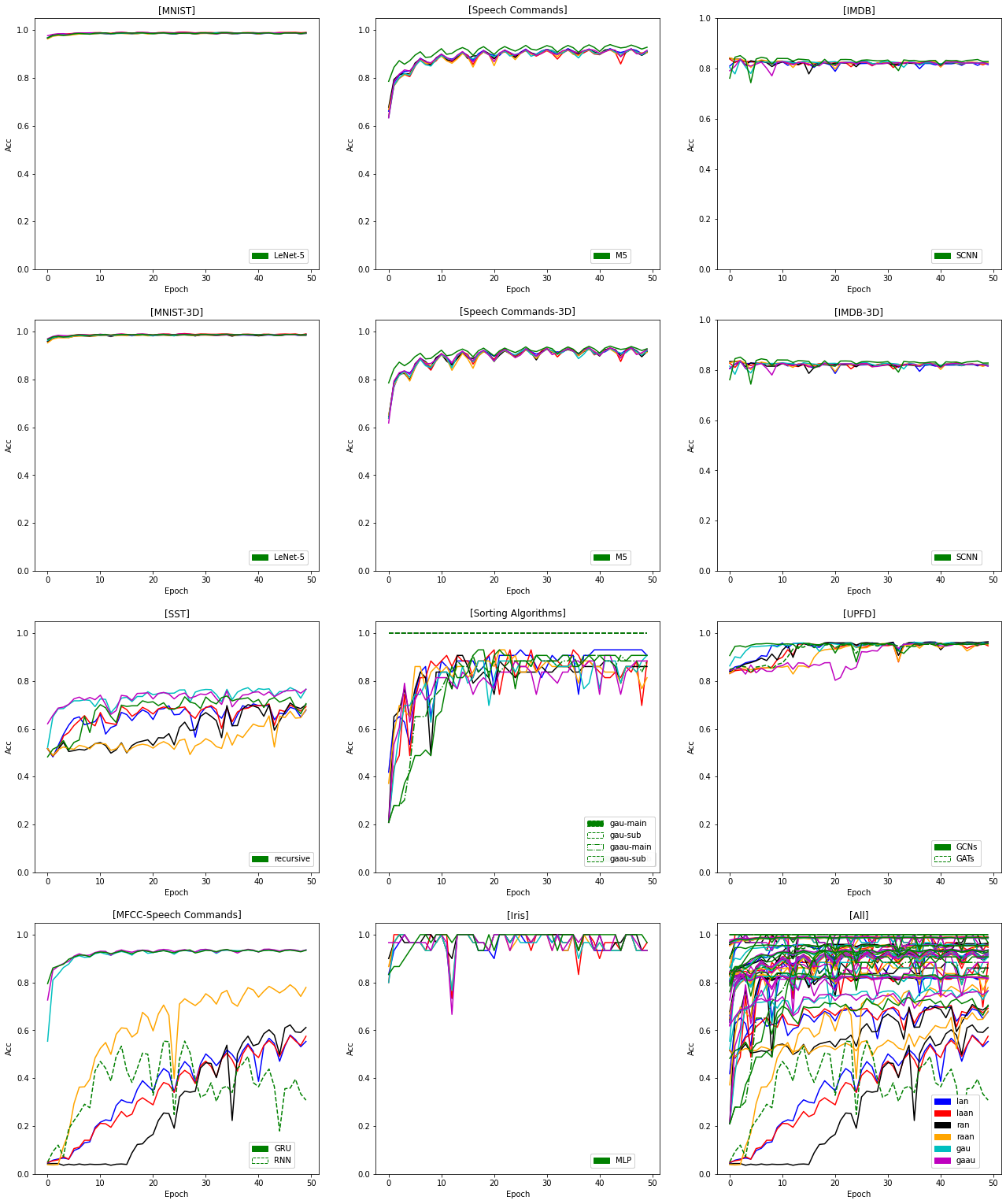}
    \hfil
    \caption{ The green line shows the test accuracy of the baseline model, and the other colors correspond to the AAN models. As shown at the bottom right, it is difficult to distinguish the plots if all plots are visualized together. Thus, we divided this subfigure into 11 plots.}
\label{fig:experimental-5-plot}
\end{figure*}
\end{document}